\pdfoutput=1

\documentclass[11pt]{article}

\usepackage[final]{acl}

\usepackage{times}
\usepackage{latexsym}
\usepackage{multirow}
\usepackage{tabularx}
\usepackage{amsmath}
\usepackage{booktabs}
\usepackage{float}
\usepackage{algorithm}
\usepackage{algorithmic}
\usepackage{amssymb}
\usepackage{pifont}
\usepackage{authblk}

\usepackage[T1]{fontenc}

\usepackage[utf8]{inputenc}

\usepackage{microtype}

\usepackage{inconsolata}

\usepackage{graphicx}
\graphicspath{{images/}}

\title{Maximum Score Routing For Mixture-of-Experts}

\author{
 \textbf{Bowen Dong\textsuperscript{1}},
 \textbf{Yilong Fan\textsuperscript{2}},
 \textbf{Yutao Sun\textsuperscript{1}},
 \textbf{Zhenyu Li\textsuperscript{1}},
\\
 \textbf{Tengyu Pan\textsuperscript{1}},
 \textbf{Xun Zhou\textsuperscript{3$\dagger$}},
 \textbf{Jianyong Wang\textsuperscript{1$\dagger$}}
\\
\\
 \textsuperscript{1}Tsinghua University,
 \textsuperscript{2}Tianjin University,
\\
 \textsuperscript{3}Seed-Foundation-Model Team, ByteDance
\\
\small{
    {dbw22@mails.tsinghua.edu.cn}
}
}

\begin{document}
\maketitle
\begin{abstract}
Routing networks in sparsely activated mixture-of-experts (MoE) dynamically allocate input tokens to top-k experts through differentiable sparse transformations, enabling scalable model capacity while preserving computational efficiency.
Traditional MoE networks impose an expert capacity constraint to ensure GPU-friendly computation.
However, this leads to token dropping when capacity is saturated and results in low hardware efficiency due to padding in underutilized experts.
Removing the capacity constraint, in turn, compromises load balancing and computational efficiency.
To address these issues, we propose Maximum Score Routing (\textbf{MaxScore}), a novel MoE routing paradigm that models routing as a minimum-cost maximum-flow problem and integrates a $\mathrm{SoftTopk}$ operator.     
MaxScore resolves the fundamental limitations of iterative rerouting and optimal transport formulations, achieving lower training losses and higher evaluation scores at equivalent FLOPs compared to both constrained and unconstrained baselines.     
Implementation details and experimental configurations can be obtained from \url{https://github.com/dongbw18/MaxScore.git}.
\end{abstract}

{\let\thefootnote\relax\footnotetext{$^\dagger$ indicates corresponding authors.}}

\section{INTRODUCTION}
The Mixture of Experts (MoE) paradigm has emerged as a compelling architectural strategy for scaling neural networks while maintaining computational efficiency. This approach dynamically combines multiple subsets of parameters (experts) by a learnable routing network, aiming to improve model capacity and computational efficiency. The routing network of sparsely activated MoE~\cite{MoE2017} dynamically allocates input tokens to top-k experts through differentiable sparse transformations, enabling conditional computation that scales model parameters without proportionally increasing FLOPs.

Softmax is conventionally employed to compute token-expert affinity coefficients in MoE routing networks, which promotes inter-expert competition. To mitigate winner-takes-all and preserve load balance, both hard constraints using expert capacity~\cite{HardConstraint}, and soft constraints using auxiliary losses~\cite{SoftConstraint}, are incorporated into the routing network~\cite{MoE2017}. 
GShard~\cite{GShard} pioneers the integration of MoE with Transformer architectures~\cite{AttentionIsAllYouNeed}, where expert capacity constraints enable GPU-friendly computation patterns. ExpertChoice~\cite{ExpertChoice} directly enables experts to select tokens based on capacity constraints. 
However, token dropping occurs when inputs are routed to capacity-saturated experts, while padding operations in underutilized experts create hardware inefficiencies. 
Empirical analysis reveals that approaches such as expanding capacity \cite{Tutel} or removing capacity constraints altogether \cite{MegaBlocks,OLMoE} effectively eliminate token dropping, but inevitably introduce a trade-off between computational efficiency and load balancing performance.
Efforts to prevent token dropping via refined routing strategies~\cite{SwitchTransformers,SBASE} have not yielded performance improvements, highlighting unresolved challenges in dynamic resource allocation.

This work introduces \textbf{Maximum Score Routing (MaxScore)}, a novel MoE routing paradigm that formulates token-expert routing as a minimum-cost maximum-flow problem~\cite{NetworkFlow}, integrated with a $\mathrm{SoftTopk}$ operator. 
To the best of our knowledge, this is the first successful integration of network flow modeling and $\mathrm{SoftTopk}$ in MoE routing.

MaxScore preserves GPU-compatible expert capacity constraints and achieves better load balancing.
Under the same FLOPs, MaxScore exhibits lower training loss and higher evaluation scores compared to both constrained and unconstrained baselines. 
Ablation studies demonstrate the necessity of both network flow modeling and the $\mathrm{SoftTopk}$ operator, revealing fundamental limitations in the iterative rerouting mechanism of ~\citet{SwitchTransformers} and the optimal transport-based routing of ~\citet{SBASE}.
The synergistic combination of two methodological enhancements yields superadditive performance gains, with empirical results demonstrating that their integrated efficacy surpasses the linear summation of individual improvements.
Scaling experiments show that MaxScore delivers consistent performance improvements with larger activated parameter budgets, and achieves more gains when increasing the number of experts, compared with standard MoE approaches.

\section{PRELIMINARIES}
\subsection{Top-k Sparsely Activated MoE}
The top-k routing mechanism is a cornerstone of sparsely activated MoE architectures, enabling efficient scaling of model capacity while maintaining computational tractability.
Originally popularized in language modeling~\cite{MoE2017}, this paradigm dynamically routes each input token to a subset of $k$ expert networks (where $k \ll e$, for $e$ total experts).
Unlike dense models that activate all parameters per input, top-k routing induces conditional computation by selecting experts based on learned gating scores, typically computed via softmax over a trainable projection of input embeddings~\cite{GShard}. 

For a given input $x$, the output $y$ of the MoE module can be written as follows:
\begin{equation}
    \label{eq:MoE}
    y = \sum_{i=1}^E R(x)_i \cdot E_i(x), \\
\end{equation}
\begin{equation}
    \label{eq:Routing}
    R(x) = \mathrm{KeepTopk}(\mathrm{Softmax}(x \cdot W_g)), \\
\end{equation}
where $R(x)$ is the sparsely activated routing function, 
$\mathrm{KeepTopk}(\cdot)$ retains the top-k largest values while setting others to zero, 
$W_g$ is the weight matrix of the routing function,
$E_i(x)$ is the output of the $i$-th expert network and the computation is performed only when $R(x)_i > 0$.

By leveraging sparse activation, MoE decouples total capacity $\mathcal{O}(e)$ from per-step computational cost, activating only $\mathcal{O}(k)$ parameters during both training and inference.

\subsection{Operators in Top-k MoE Routing\label{sec22}}
Routings in MoE commonly use $\mathrm{Softmax}(\cdot)$ to calculate the token-expert affinity coefficients, which encourages competition between experts.
However, $\mathrm{Softmax}(\cdot)$ serves as a smooth approximation to the one-hot $\mathrm{Argmax}(\cdot)$ function, which can lead to inefficiencies in top-$k$ routing, as the top-1 expert often receives a disproportionately large affinity score compared to the remaining $k{-}1$ experts.

Alternative routing operators have also been investigated. 
\citet{DeepSeekV3} replaces $\mathrm{Softmax}(\cdot)$ with $\mathrm{Sigmoid}(\cdot)$ to align with its auxiliary-loss-free load balancing strategy, while ReMoE~\cite{ReMoE} explores the feasibility of using $\mathrm{ReLU}(\cdot)$ for routing decisions.

We define $\mathrm{SoftTopk}(\cdot)$ as a smooth approximation to $\mathrm{ArgTopk}(\cdot)$, which represents the top-k selection in a one-hot form, formally given by:
\begin{equation}
    \mathrm{ArgTopk}(\mathbf{a})_i = \begin{cases}
        1, a_i \in \mathrm{Topk}(\mathbf{a})\\
        0, \mathrm{otherwise},\\
    \end{cases}
\end{equation}
where $\mathbf{a}=(a_1,a_2,...,a_e)$ represents the affinity coefficients between the token and e experts.

\citet{SparseMax} and \citet{EntMax} proposed $\mathrm{Sparsemax}(\cdot)$ and $\mathrm{Entmax}(\cdot)$ as differentiable approximations for top-k probability truncation.
\citet{SoftTopk} further introduced a broader family of $\mathrm{SoftTopk}(\cdot)$ operators. 
However, their integration into MoE routing has not been investigated, leaving a promising direction underexplored.


\begin{figure}[t]
  \centering
  \setlength{\abovecaptionskip}{0pt}
  \includegraphics[width=0.98\linewidth]{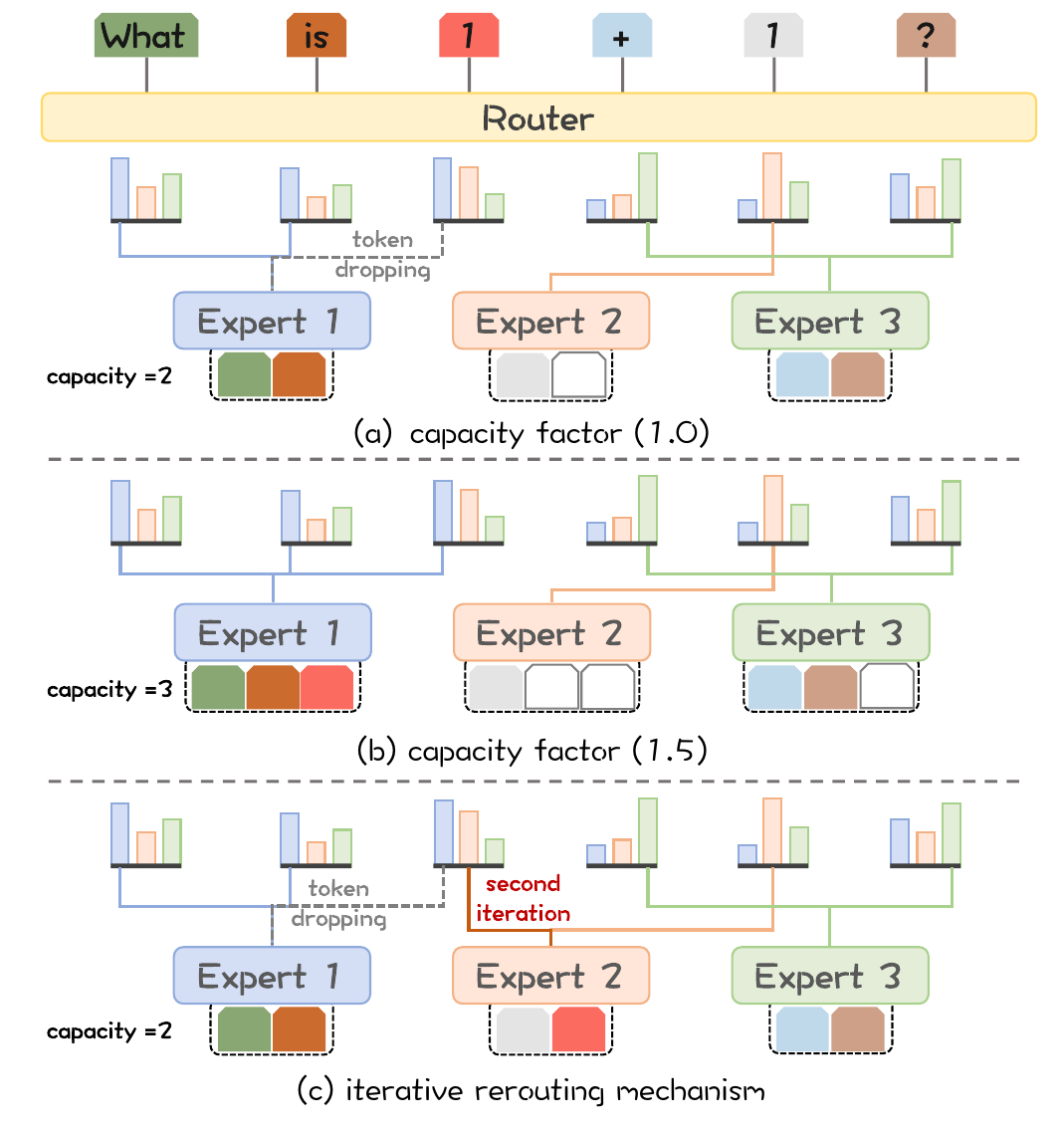} 
  \caption {\label{newFig1:ExperCapacity}
    Different top-2 routing paradigms for 3 experts and 6 tokens. 
      (a) sets capacity-factor $c_f = 1.0$,  and token dropping occurs;
      (b) sets capacity-factor $c_f = 1.5$,  there is no more token dropping, but more computation is wasted;
      (c) uses iterative rerouting mechanism, the dropped token is reassigned to expert with remaining capacity.
  }\vspace{-10pt}
\end{figure}

\subsection{Expert Capacity Constrained}
To counteract the winner-takes-all phenomenon and maintain load balancing in the routing network, traditional routing architectures integrate dual constraint mechanisms: (i) hard limits through expert capacity~\cite{HardConstraint}, and (ii) soft regularization via differentiable auxiliary losses~\cite{SoftConstraint,MoE2017,Router-z-Loss}. 

\begin{figure}[t]
  \centering
  \setlength{\abovecaptionskip}{0pt}
  \includegraphics[width=\linewidth]{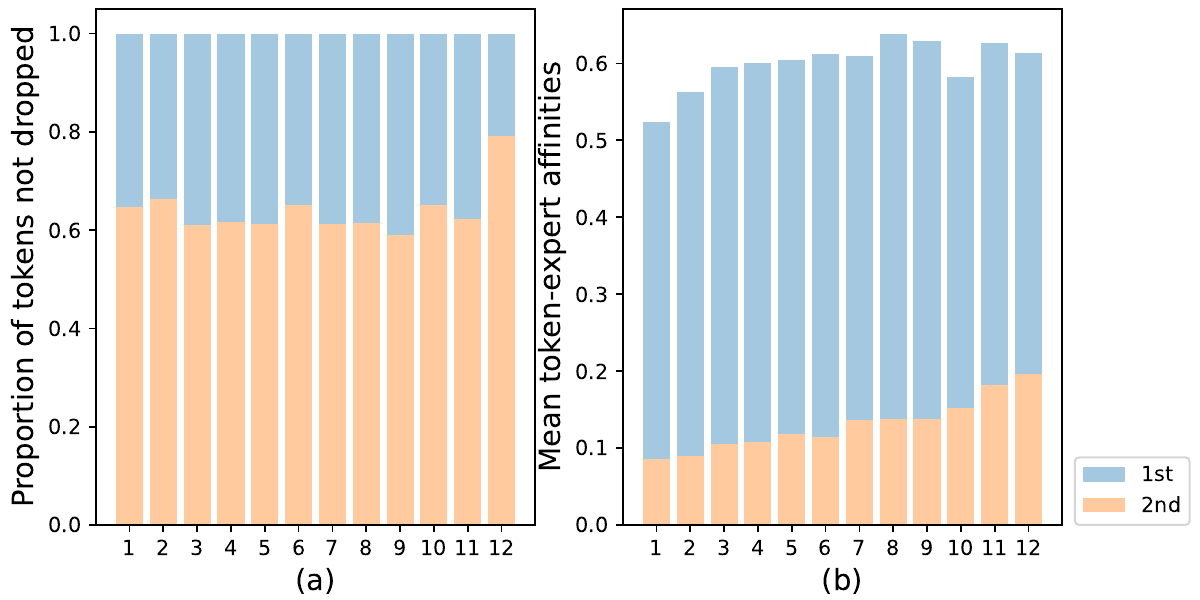}
  \caption {\label{newFig2:ratio}
    The proportion of tokens not dropped and the mean token-expert affinities in top-2 routing are analyzed separately. 
    The data is derived from the GShard MoE with $e=16$ after training on $65$ billion tokens. 
    (a) shows that tokens assigned to the top-1 experts are rarely dropped, whereas approximately $35\%$ of tokens routed to the second experts experience dropping. 
    (b) illustrates that the top-1 token-expert affinities are typically much higher than those of other experts.
  }\vspace{-10pt}
\end{figure}

GShard~\cite{GShard} strategically harmonizes capacity-constrained MoE design with Transformer architectures~\cite{AttentionIsAllYouNeed}.
For a batch of $n$ tokens, GShard fixes per-expert capacity with $c = \frac{k * n}{e}$ to enable parallel-friendly computation patterns.
This routing mechanism, however, poses optimization challenges due to imbalanced expert utilization. 
While underloaded experts incur computational overhead through padding (mathematically sound but hardware-inefficient), overloaded experts lead to token dropping.
Increasing expert capacity $c' = c_f * \frac{k * n}{e}$ by a capacity-factor $c_f$ can alleviate token dropping. 
Tutel~\cite{Tutel} uses a highly scalable stack design and sets the $c_f$ dynamically, but it would lead to additional computational costs and reduced load balancing.
Figure~\ref{newFig1:ExperCapacity}(a) and ~\ref{newFig1:ExperCapacity}(b) shows the trade-off between token dropping and additional computation by increasing expert capacity.
Figure~\ref{newFig2:ratio}(a) shows the token dropping proportion in the MoE routing of each layer in a GShard model with $e=16$ and $k=2$, and approximately $35\%$ of tokens routed to the second experts experience dropping.

ExpertChoice~\cite{ExpertChoice} inverts the conventional routing paradigm by allowing experts to select their top-c tokens, thereby achieving optimal load balancing. 
However, this strategy allows each token to be assigned to an arbitrary number of experts, including zero, which exacerbates token dropping. 
More importantly, it introduces a data leakage issue: determining whether a token belongs to the top-c set of a given expert requires comparisons not only with preceding tokens but also with subsequent ones, thereby violating the causal structure required by autoregressive models.

Another class of approaches, referred to as DropLess MoE, eliminates capacity constraints entirely to prevent token dropping. Those methods allocate an indefinite number of tokens to experts via direct indexing (e.g., DeepSeekMoE~\cite{DeepSeekV1,DeepSeekV2,DeepSeekV3}, OLMoE~\cite{OLMoE,MegaBlocks}).

Switch Transformers~\cite{SwitchTransformers} explored an iterative rerouting mechanism for dropped tokens as shown in Figure~\ref{newFig1:ExperCapacity}(c): in the first stage, tokens are assigned to experts using the top-k strategy; in the second stage, any dropped tokens are greedily reassigned to the highest-affinity expert among those with remaining capacity. However, empirical results show that this approach does not lead to improvements in model quality.

SBASE~\cite{SBASE} formulates MoE routing as an optimal transport problem: $\mathbf{c}=(c_1,c_2,...,c_e)$ denotes the capacity of each expert, and $\mathbf{k}=(k_1,k_2,...,k_n)$ specifies the number of experts each token should be assigned to. The matrix $\mathbf{A}\in \mathbb{R}^{n\times e}$ represents token-expert affinity coefficients. The feasible solution space is defined as 
\begin{equation}
U(\mathbf{c},\mathbf{k}) = \{\mathbf{P} \in \mathbb{R}^{n \times e}_{\geq0}|\mathbf{P}^T\mathbf{1}_n=\mathbf{c},\mathbf{P}\mathbf{1}_e=\mathbf{k}\},
\label{eq4}
\end{equation}
and the optimization objective is 
\begin{equation}
d_{\mathbf{A}}(\mathbf{c}, \mathbf{k}) = \max_{\mathbf{P} \in U(\mathbf{c}, \mathbf{k})} \sum_{ij} \mathbf{P}_{ij} \mathbf{A}_{ij}.
\label{eq5}
\end{equation}
To efficiently approximate the solution, SBASE employs the parallelizable Sinkhorn algorithm~\cite{Sinkhorn}. Nonetheless, this formulation primarily contributes to improved training stability, offering limited gains beyond this benefit.

\section{METHODOLOGY}
We investigate the fundamental reasons why the iterative rerouting mechanism (\textbf{Iter}) and the optimal transport formulation (\textbf{Sinkhorn}) fail to improve model quality, and propose Maximum Score Routing (\textbf{MaxScore}), a novel mixture-of-experts routing strategy that integrates network flow modeling and a differentiable $\mathrm{SoftTopk}(\cdot)$ operator.

\subsection{Limitations of Iter and Sinkhorn}
\noindent{\textbf{Softmax operator}.}
Both the iterative rerouting mechanism and the optimal transport formulation aim to achieve a globally improved allocation by replacing locally optimal assignment strategies.  
However, as discussed in Section~\ref{sec22}, using the conventional $\mathrm{Softmax}(\cdot)$ to compute token-expert affinity scores results in the top-1 affinity being significantly higher than those of other token-expert pairs.  
We statistically analyze the probability distribution in a top-2 GShard MoE, as shown in Figure~\ref{newFig2:ratio}(b), where the top-1 token-expert affinities markedly exceeds that of the second-ranked expert.  
For example, if a token’s top-2 affinities are $0.8$ and $0.05$ respectively, then when the first expert is saturated, substituting with any expert outside the top-2 (with affinity below $0.05$) yields no meaningful benefit;  similarly, if the second expert is saturated, replacing it has negligible impact on the model’s gradient.

\begin{figure}[ht]
  \centering
  \setlength{\abovecaptionskip}{0pt}
  \includegraphics[width=0.9\linewidth]{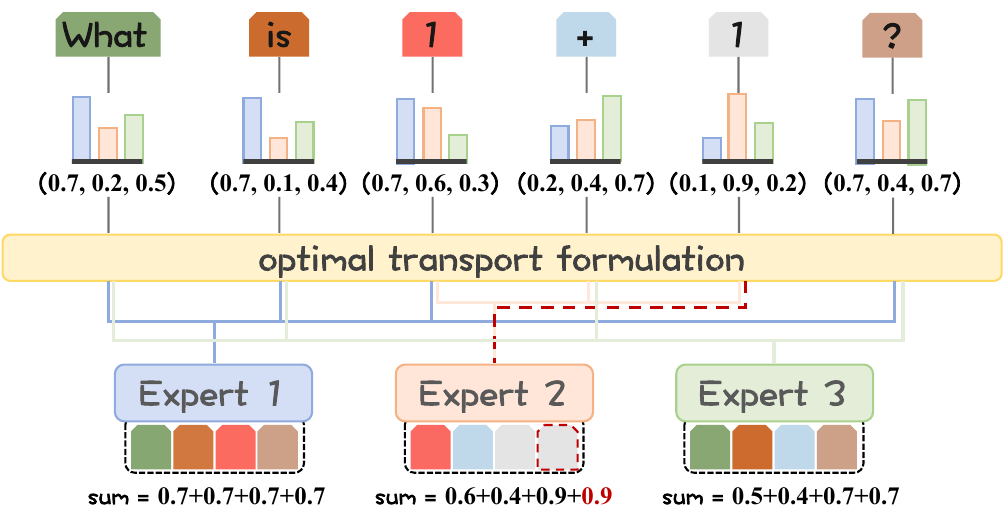} 
  \caption {\label{newFig3:Sinkhorn}
    Limitation of optimal transport formulation.
    The fifth token and the second expert matched twice.
  } \vspace{-10pt}
\end{figure}

\noindent{\textbf{Limitation of optimal transport formulation}.}
Modeling MoE routing using Equations~(\ref{eq4}) and~(\ref{eq5}) has inherent limitations: in MoE routing strategies, the actual gain of a token-expert pair appearing multiple times is equivalent to that of a single occurrence.  
This constraint cannot be enforced in the optimal transport formulation.  
As illustrated in Figure~\ref{newFig3:Sinkhorn}, high-probability token-expert pairs may be matched repeatedly, causing redundant reward accumulation and effectively degenerating to a top-1 routing scheme, which results in wasted computational resources.

\subsection{Maximum Score Routing}

\noindent{\textbf{SoftTopk operator}.}
We first tried different operators as shown in Table~\ref{Tab2:softtopk}, but due to the potential damage caused by the increased computational complexity, we did not achieve better results than $\mathrm{Softmax}(\cdot)$.
We propose a simple but highly effective $\mathrm{SoftTopk}(\cdot)$ operator for MoE routing:
\begin{equation}
        \begin{aligned}
        \mathrm{SoftTopk}(\mathbf{a})^{(k)} &= \mathrm{SoftTopk}(\mathbf{a})^{(k-1)} + \mathrm{SE}(\mathbf{a}), \\
            \mathrm{SE}(\mathbf{a})_i &= \begin{cases}
                0, a_i \in \mathrm{Topk}(\mathbf{a})\\
                t \cdot \mathrm{Softmax}(\mathbf{a})_i, \mathrm{otherwise},\\
            \end{cases}
        \end{aligned}
    \label{Eq:SoftTopk}
\end{equation}
where $t$ is a constant that gradually decays from the initialization value $t_0$ to $0$.

\begin{table}[t]
  \centering
  \small
  \renewcommand{\arraystretch}{2}
  \begin{tabularx}{\linewidth}{ll} \toprule[1pt] \textbf{Name} & \textbf{Expression} \\
    \midrule[1pt]
    $\mathrm{Softmax}(x)$  & $y=e^{x}/\sum_j^N{e^{x_j}}$           \\
    \hline
    $\mathrm{Sigmoid}(x)$  & $y=1/(1+e^{-x})$          \\
    \hline
    \multirow{2}{*}{$\mathrm{SoftKmax}(x)^{(k)}$}  & $y^{(k)}=y^{(k-1)} + \mathrm{Softmax}(g^{(k-1)})$           \\
    & $g^{(k-1)}=(1-y^{(k-1)}) \otimes x$\\
    \hline
    \multirow{2}{*}{$\mathrm{IterTopk}(x)^{(k)}$}     & $y^{(k)}=y^{(k-1)} + g(x;1-y^{(k-1)})$           \\
    & $g(x;w)=w\cdot e^{x}/\sum_j^N{w_j\cdot e^{x_j}}$ \\
    \hline
    \multirow{3}{*}{$\mathrm{GradTopk}(x)^{(k)}$} & $y^{(k)} = e^{g^{(k)}-z^{(k)}}$\\
    & $g^{(k)} = x + log(e^{z^{(k-1)}}-e^{g^{(k-1)}})$\\
    & $z^{(k)} = log(\sum_j^Ne^{g^{(k)}_j}) - logk$ \\
    \bottomrule[1pt]
  \end{tabularx}
  \caption{\label{Tab2:softtopk}
    Operators can be used for MoE routing. $\mathrm{SoftKmax}$, $\mathrm{IterTopk}$ and $\mathrm{GradTopk}$ are mentioned in \citet{SoftTopk}.
  }
\end{table}

\begin{figure}[ht]
  \centering
  \setlength{\abovecaptionskip}{0pt}
  \includegraphics[width=0.9\linewidth]{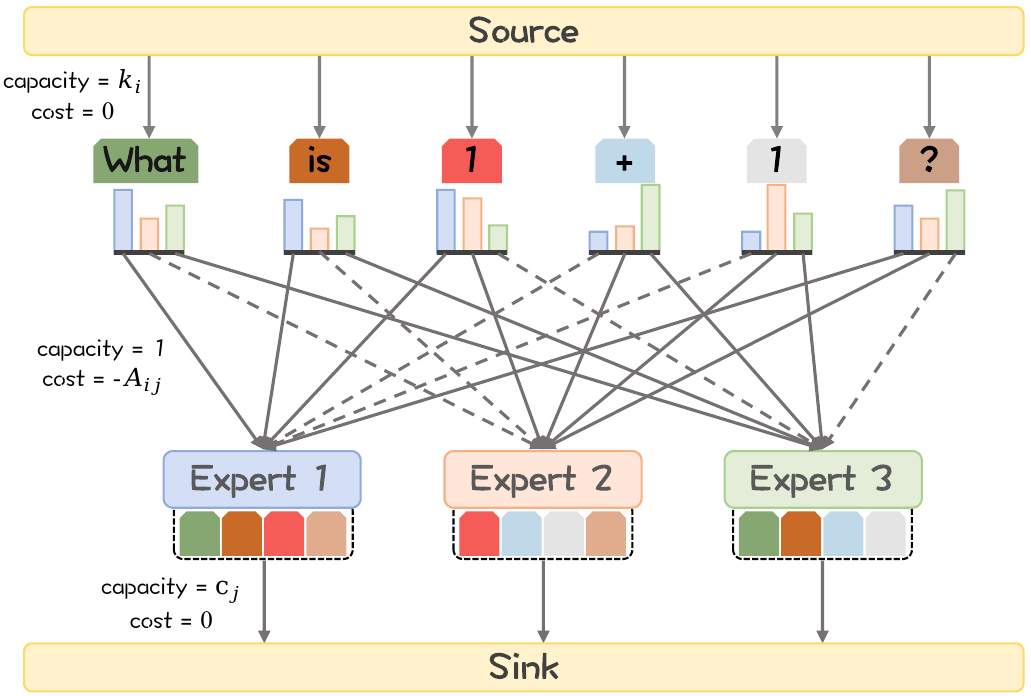} 
  \caption {\label{newFig4:NetworkFlow}
    The minimum-cost maximum-flow modeling for MoE routing.
  } \vspace{-10pt}
\end{figure}

\noindent{\textbf{Network flow modeling}.}
To better capture the characteristics of MoE routing, Equations~(\ref{eq4}) and~(\ref{eq5}) are revised as follows:
\begin{equation}
U'(\mathbf{c},\mathbf{k}) = \{\mathbf{P} \in \mathbb{F}_{2}^{n \times e}|\mathbf{P}^T\mathbf{1}_n=\mathbf{c},\mathbf{P}\mathbf{1}_e=\mathbf{k}\},
\label{eq7}
\end{equation}
\begin{equation}
d'_{\mathbf{A}}(\mathbf{c}, \mathbf{k}) = \max_{\mathbf{P} \in U'(\mathbf{c}, \mathbf{k})} \sum_{ij} \mathbf{P}_{ij} \mathbf{A}_{ij},
\label{eq8}
\end{equation}
where $\mathbb{F}_{2}$ denotes the finite field of $\{0, 1\}$ equipped with addition and multiplication operations.
To address this problem, MoE routing can be formulated as a minimum-cost maximum-flow problem as shown in Figure~\ref{newFig4:NetworkFlow}.
We model tokens and experts as nodes in a flow network graph. 
Edges from the super source to tokens have capacities representing that each token must be assigned to k experts, while edges from experts to the super sink enforce capacity constraints of c per expert. 
These edges carry zero cost. 
Edges between tokens and experts have unit capacity, allowing at most one match per token-expert pair, with costs defined as the negation of their affinity coefficients. 
A detailed summary of the graph edge properties is provided in Table~\ref{tab1:Edges}.

\begin{table}[ht]
  \centering
  \small
  \renewcommand{\arraystretch}{1.3}
  \begin{tabularx}{0.9\linewidth}{ccccc} \toprule[1pt] \textbf{From} & \textbf{To} & \textbf{Capacity} & \textbf{Cost} & \textbf{Count} \\
    \midrule[1pt]
    $\mathrm{Source}$ & $\mathrm{Token}_i$ & $k_i$ & $0$ & $n$\\
    $\mathrm{Expert}_j$ & $\mathrm{Sink}$ & $c_j$ & $0$ & $e$\\
    $\mathrm{Token}_i$ & $\mathrm{Expert}_j$ & $1$ & $-A_{ij}$ & $n * e$\\ \bottomrule[1pt]
  \end{tabularx}
  \caption{\label{tab1:Edges}
    Edges in the graph of Figure~\ref{newFig4:NetworkFlow}. 
    $\mathrm{Source}$ is the super source, $\mathrm{Sink}$ is the super sink, $A_{ij}$ represents the affinity coefficient between $\mathrm{Token}_i$ and $\mathrm{Expert}_j$.
  }
\end{table}

\noindent{\textbf{Algorithm complexity optimization}.} 
TA commonly used and effective approach to solving the minimum-cost maximum-flow problem is the Shortest Path Faster Algorithm (SPFA)~\cite{Bellman,Ford}, which iteratively searches for the lowest-cost augmenting path until no such path remains. 
However, this method is computationally expensive and inherently sequential, limiting its parallelizability. 
In top-2 MoE routing, given that the token drop rate in top-1 routing is relatively low (approximately $0$ as shown in Figure~\ref{newFig2:ratio}) and that the Sinkhorn algorithm corresponds to the minimum-cost maximum-flow formulation under top-1 routing, we propose a two-stage strategy: first allocate tokens using top-1 routing, followed by applying the Sinkhorn algorithm to handle the residual routing problem.
The complete algorithm process is shown in Algorithm~\ref{Algorithm}.
For top-k MoE routing with $k>2$, a trade-off needs to be made between quality (SPFA) and speed (Iter).

\begin{table*}[t]
  \centering
  \small
  \renewcommand{\arraystretch}{1.4}
 \begin{tabularx}{\linewidth}{lccccccccccc} \toprule[1pt]
      \multirow{2}{*}{\textbf{Model}} & ARC & ARC & \multirow{2}{*}{BoolQ} & Hella- & LAM- & \multirow{2}{*}{PIQA} & \multirow{2}{*}{RACE} & \multirow{2}{*}{SciQ} & \multirow{2}{*}{Record} & \multirow{2}{*}{OBQA} & \multirow{2}{*}{Avg.}  \\ 
     & challenge & easy &  & Swag & BADA & & & & &  \\ \midrule[1pt]
    \textbf{Dense}         &  $18.69$ & $40.19$ & $57.06$ & $28.91$ & $16.28$ & $63.71$ & $25.65$ & $64.2$ & $56.05$ & $15.0$ & $38.57$ \\
    \textbf{GShard}        &  $18.86$ & $\textbf{44.49}$ & $61.90$ & $31.74$ & $21.54$ & $66.38$ & $\textbf{28.52}$ & $69.4$ & $62.08$ & $16.2$ & $42.11$ \\
    \textbf{GShard-I}      &  $19.80$ & $44.36$ & $59.94$ & $32.54$ & $21.52$ & $67.03$ & $28.23$ & $68.7$ & $62.84$ & $16.0$ & $42.10$ \\
    \textbf{SBASE}         &  $18.34$ & $43.73$ & $57.61$ & $30.96$ & $19.70$ & $65.18$ & $27.37$ & $68.3$ & $60.06$ & $16.2$ & $40.75$ \\ 
    \textbf{ExpertChoice}  &  $19.37$ & $42.00$ & $61.74$ & $32.10$ & $21.19$ & $66.16$ & $27.18$ & $68.4$ & $62.26$ & $17.6$ & $41.80$ \\
    \textbf{DropLess}      &  $19.28$ & $44.07$ & $61.16$ & $32.03$ & $21.35$ & $67.14$ & $27.08$ & $67.9$ & $61.55$ & $16.0$ & $41.76$ \\
    \textbf{DeepSeek}      &  $19.88$ & $44.28$ & $60.55$ & $32.23$ & $21.93$ & $66.97$ & $27.94$ & $70.9$ & $62.57$ & $17.6$ & $42.49$ \\ \hline
    \textbf{MaxScore-I}    &  $\textbf{20.90}$ & $43.22$ & $61.71$ & $32.51$ & $21.66$ & $\textbf{67.41}$ & $28.42$ & $69.9$ & $63.61$ & $\textbf{18.4}$ & $42.77$ \\
    \textbf{MaxScore}      &  $20.73$ & $\textbf{44.49}$ & $\textbf{62.23}$ & $\textbf{32.85}$ & $\textbf{23.27}$ & $\textbf{67.41}$ & $\textbf{28.52}$ & $\textbf{72.5}$ & $\textbf{64.00}$ & $\textbf{18.4}$ & $\textbf{43.44}$ \\ \bottomrule[1pt]
\end{tabularx}
  \caption{\label{Tab2:MainResult}
    Results for the base-sized models.
  }
\end{table*}

\begin{algorithm}
	\renewcommand{\algorithmicrequire}{\textbf{Input:}}
	\renewcommand{\algorithmicensure}{\textbf{Output:}}
	\caption{Maximum Score Routing For Top-$2$ Mixture-of-Experts}
	\label{Algorithm}
	\begin{algorithmic}[1]
            \REQUIRE Weight matrix $W_g$ in the routing function, the number of experts $e$, temperature $t \leftarrow t_0$, a batch of $n$ tokens $\{x_i\}$
            \STATE Initialization expert capacity $\mathbf{c}$: $c_j \leftarrow 2 * n / e$
            \STATE Calculate the token-expert affinity coefficients: $a_{i,j} \leftarrow \mathrm{SoftTopk}(x_i \cdot W_g)_j$
            \STATE Update temperature: $t$
            \STATE Calculate the mask matrix of top-1: $\mathrm{mask}_{i,j}\leftarrow\mathrm{onehot}(\mathrm{Argmax}(a_i),e)_j$
            \STATE Remove top-1: $a_{i,j} \leftarrow a_{i,j} \cdot \neg \mathrm{mask}_{i,j}$
            \STATE Update expert capacity $\mathbf{c}$: $c_j \leftarrow max(0,c_j - \sum_i\mathrm{mask}_{i,j})$
            \STATE Set $\mathbf{k}$: $k_i \leftarrow 1$
            \STATE The feasible solution space: $U'(\mathbf{c},\mathbf{k}) = \{\mathbf{P} \in \mathbb{F}_{2}^{n \times e}|\mathbf{P}^T\mathbf{1}_n=\mathbf{c},\mathbf{P}\mathbf{1}_e=\mathbf{k}\}$,
            \STATE Use $\mathrm{\textbf{Sinkhorn}}$ for an approximate solution: $d'_{\mathbf{A}}(\mathbf{c}, \mathbf{k}) = \max_{\mathbf{P} \in U'(\mathbf{c}, \mathbf{k})} \sum_{ij} \mathbf{P}_{ij} \mathbf{A}_{ij}$
            \ENSURE $\{\mathbf{P}_{ij}\}$
	\end{algorithmic}  
\end{algorithm}

\section{EVALUATION}

\subsection{Experimental Setup}

\noindent{\textbf{Model Architecture.}}
We conduct our experiments using the Llama architecture~\cite{LLama,LLama2,LLama3}, incorporating grouped query attention (GQA) ~\cite{GQA}, SwiGLU activation function~\cite{GLU}, RoPE position embedding~\cite{RoPE}, and RMSNorm~\cite{RMSNorm}. 
Our sparsely activated models are constructed by substituting the MLP layers of the dense baseline with MoE layers.
We explore three different backbone sizes, as detailed in Table~\ref{tab0:configs}.

\noindent{\textbf{Baselines.}}
We compared the dense model, GShard MoE~\cite{GShard} and GShard-I MoE, the variant with iterative routing strategy~\cite{SwitchTransformers}, SBASE MoE~\cite{SBASE}, ExpertChoice MoE~\cite{ExpertChoice}, DropLess MoE~\cite{MegaBlocks}, DeepSeek-V2 MoE~\cite{DeepSeekV1,DeepSeekV2} along with our proposed \textbf{MaxScore} MoE and \textbf{MaxScore-I} MoE, which replaces network flow modeling with the iterative rerouting mechanism. 
All MoEs except DeepSeek use the base configuration with $k=2$ and $e=16$, while DeepSeek MoE employs fine-grained experts with $k=6$ and $e=64$ and a double-sized shared expert.

\noindent{\textbf{Load Balance Loss.}}
All MoE models employ the same auxiliary loss function, defined as
\begin{equation}
\mathcal{L}_{\mathrm{aux}} = \lambda \cdot \frac{1}{e} \sum_{j=1}^e \left( \frac{1}{n} \sum_{i=1}^n \mathbf{A}_{i,j} \right) \left( \frac{1}{n} \sum_{i=1}^N \mathbf{P}_{i,j} \right),
\label{eq9}
\end{equation}
where the $\mathbf{A}_{i,j}$ and $\mathbf{P}_{i,j}$ correspond to the terms defined in Equations~(\ref{eq7}) and~(\ref{eq8}). 

\noindent{\textbf{Training Settings.}}
We adopt the tokenizer from LLama~\cite{LLama,LLama2,LLama3} and set the context length to 512.  
The batch size is 688, which is the largest setting that allows all baseline models to be trained on 8 NVIDIA A800 GPUs (this constraint arises primarily from the DeepSeek, as shown in Table~\ref{tab3:GPU}).
We can train all baselines with $8$ NVIDIA A800 GPUs. 
All models are trained for $180 k$ steps (approximately $65 B$ tokens) on C4 dataset~\cite{2019t5}.
This exceeds the compute-optimal dataset size identified by \citet{scalinglaw}, ensuring convergence.
For training, we leverage the HuggingFace Trainer~\cite{huggingface} integrated with DeepSpeed optimizations, including Zero Redundancy Optimizer (ZeRO)~\cite{ZeRO} and activation checkpointing~\cite{activation_ckpt}, and we employ bfloat16 for numerical precision and efficiency.
We adopt AdamW~\cite{AdamW} as the optimizer with weight decay $wd$, adam betas $\mathrm{(\beta_1,\beta_2)}$ and adam epsilon $\mathrm{\epsilon}$.
The learning rate is set to be $lr$ following a WSD scheduler~\cite{WSD} with a warmup for $2k$ steps and decay over the last $6k$ steps. 

\noindent{\textbf{Hyperparameters.}}
We perform grid searchs over learning rate $lr$, weight decay $wd$, adam betas $\mathrm{(\beta_1,\beta_2)}$, and adam epsilon $\mathrm{\epsilon}$ on the GShard baseline, and apply the selected hyperparameters uniformly across all other baselines, as summarized in Table~\ref{tab5:hyperparameters}.
For the scaling factor $\lambda$ of the auxiliary loss in Equation~(\ref{eq9}), we perform a grid search over the set $\{10^{-1}, 10^{-2}, 10^{-3}, 10^{-4}\}$ for each baseline. 
The final selected values are $10^{-3}$ for DeepSeek and $10^{-2}$ for all other baselines.

\noindent{\textbf{Evaluation Settings.}}
We leverage the open source lm-evaluation-harness~\cite{lm-eval} for standardized evaluation on various types of tasks:
ARC challenge, ARC easy~\cite{ARC}, 
BoolQ~\cite{BoolQ},
HellaSwag~\cite{HellaSwag},
LAMBADA~\cite{LAMBADA},
PIQA~\cite{PIQA},
RACE~\cite{RACE},
SciQ~\cite{SciQ},
Record~\cite{ReCoRD} and OpenBookQA (OBQA)~\cite{OpenBookQA}.


\begin{figure}[t]
  \setlength{\abovecaptionskip}{4pt}
  \includegraphics[width=\linewidth]{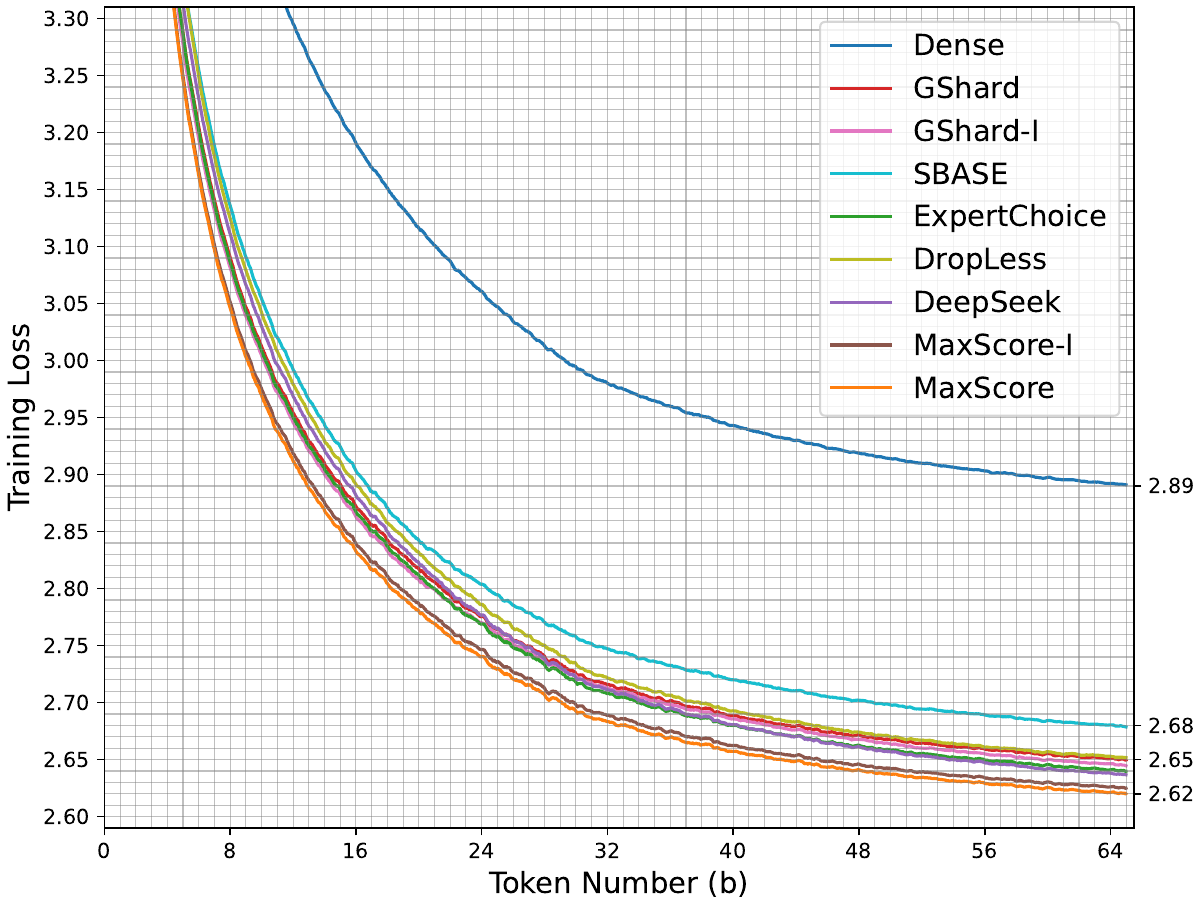}
  \caption {\label{Fig4:MainResult}
  Training loss curve.
  } 
\end{figure}

\begin{table*}[ht]
  \centering
  \scriptsize
  \renewcommand{\arraystretch}{1.4}
 \begin{tabularx}{0.97\linewidth}{lccccccccccc} \toprule[1pt]
      \multirow{2}{*}{\textbf{Model}} & ARC & ARC & \multirow{2}{*}{BoolQ} & Hella- & LAM- & \multirow{2}{*}{PIQA} & \multirow{2}{*}{RACE} & \multirow{2}{*}{SciQ} & \multirow{2}{*}{Record} & \multirow{2}{*}{OBQA} & \multirow{2}{*}{Avg.}  \\ 
     & challenge & easy &  & Swag & BADA & & & & &  \\ \midrule[1pt]
    \textbf{GShard}                 &  $18.86$ & $\mathbf{44.49}$ & $61.90$ & $31.74$ & $21.54$ & $66.38$ & $\mathbf{28.52}$ & $69.4$ & $62.08$ & $16.2$ & $42.11$ \\
    \textbf{GShard-I}               &  $19.80$ & $44.36$ & $59.94$ & $32.54$ & $21.52$ & $67.03$ & $28.23$ & $68.7$ & $62.84$ & $16.0$ & $42.10$ \\
    \textbf{GShard-M}               &  $20.14$ & $43.74$ & $59.38$ & $32.27$ & $22.30$ & $66.63$ & $27.61$ & $68.7$ & $62.59$ & $18.2$ & $42.16$ \\
    \textbf{GShard-S}               &  $20.52$ & $44.30$ & $59.13$ & $32.34$ & $22.54$ & $66.74$ & $28.19$ & $69.4$ & $63.94$ & $\mathbf{18.4}$ & $42.55$ \\
    \textbf{GShard-SI (MaxScore-I)} &  $\mathbf{20.90}$ & $43.22$ & $61.71$ & $32.51$ & $21.66$ & $\mathbf{67.41}$ & $28.42$ & $69.9$ & $63.61$ & $\mathbf{18.4}$ & $42.77$ \\    \textbf{GShard-SM (MaxScore)}   &  $20.73$ & $\mathbf{44.49}$ & $\mathbf{62.23}$ & $\mathbf{32.85}$ & $\mathbf{23.27}$ & $\mathbf{67.41}$ & $\mathbf{28.52}$ & $\mathbf{72.5}$ & $\mathbf{64.00}$ & $\mathbf{18.4}$ & $\mathbf{43.44}$ \\ \bottomrule[1pt]
\end{tabularx}
  \caption{\label{Tab3:Ablation}
    Ablation study results. 
    We validate the contributions of the $\mathrm{SoftTopk(\cdot)}$ Operator (\textbf{S}), the Minimum-cost Maximum Flow Modeling (\textbf{M}), and the Iterative Routing Strategy (\textbf{I}).
  }
\end{table*}

\subsection{Main Results}
Figure~\ref{Fig4:MainResult} presents the training loss curves for all evaluated base-sized models, and Table~\ref{Tab2:MainResult} summarizes the evaluation results of models after training on about $65B$ tokens.

Our proposed MaxScore and MaxScore-I consistently achieve lower training loss compared to all baseline methods throughout the training process and outperform existing baselines on the evaluation datasets.
Notably, MaxScore attains the lowest final training loss of approximately $2.62$, indicating more effective optimization and improved convergence behavior, and achieves the highest average accuracy of $43.44\%$, surpassing the best baseline (DeepSeek) by approximately $0.95\%$.
It also attains state-of-the-art performance on almost all individual tasks.
The iterative variant MaxScore-I demonstrates competitive results, particularly excelling on ARC challenge and PIQA.

These findings validate the superiority of our routing mechanisms in integrating the $\mathrm{SoftTopk(\cdot)}$ operator and the minimum cost maximum flow modeling in improving MoE routing quality.

\begin{table}[ht]
  \centering
  \small
  \renewcommand{\arraystretch}{1.3}
  \begin{tabular}{ccc} \toprule[1pt] \textbf{Name} & \textbf{Gird Search} & \textbf{Result} \\
    \midrule[1pt]
    $lr$ & $\{\{1,3\}*\{10^{-4}, 10^{-5}, 10^{-6} \}\}$ & $3 * 10^{-5} $ \\
    $wd$ & $\{\{0,1,2,3,4\}*0.05\}$ & $0.1$ \\
    $\mathrm{(\beta_1,\beta_2)}$ & $(0.9, \{0.999,0.99,0.95,0.9\}) $ & $(0.9, 0.95)$ \\
    $\mathrm{\epsilon}$ & $\{10^{-5}, 10^{-6}, 10^{-7}, 10^{-8}\}$ & $10^{-6}$ \\
    \bottomrule[1pt]
  \end{tabular}
  \caption{\label{tab5:hyperparameters}
    Gird search and results for hyperparameters.
  }
\vspace{-10pt}
\end{table}

\subsection{Ablation Evaluation}

Table~\ref{Tab3:Ablation} presents the ablation study results, validating the individual contributions of the $\mathrm{SoftTopk(\cdot)}$ operator (\textbf{S}), the minimum-cost maximum flow modeling (\textbf{M}), and the iterative routing strategy (\textbf{I}).
The variants GShard-S, GShard-M, and GShard-I correspond to incorporating SoftTopk, network flow modeling, and iterative routing respectively, while GShard-SI (MaxScore-I) and GShard-SM (MaxScore) combine these components.

GShard exhibits negligible improvements when employing either network flow modeling or the iterative strategy alone, consistent with observations reported in SwitchTransformer.       However, incorporating the $\mathrm{SoftTopk(\cdot)}$ operator individually yields noticeable gains.       
Furthermore, combining the iterative strategy or network flow modeling with the $\mathrm{SoftTopk(\cdot)}$ operator results in substantial performance improvements.
This demonstrates the necessity of the $\mathrm{SoftTopk(\cdot)}$ operator, revealing fundamental limitations in the iterative rerouting mechanism of ~\citet{SwitchTransformers} and the optimal transport-based routing of ~\citet{SBASE}.

\begin{figure}[t]
  \centering
  \setlength{\abovecaptionskip}{4pt}
  \includegraphics[width=\linewidth]{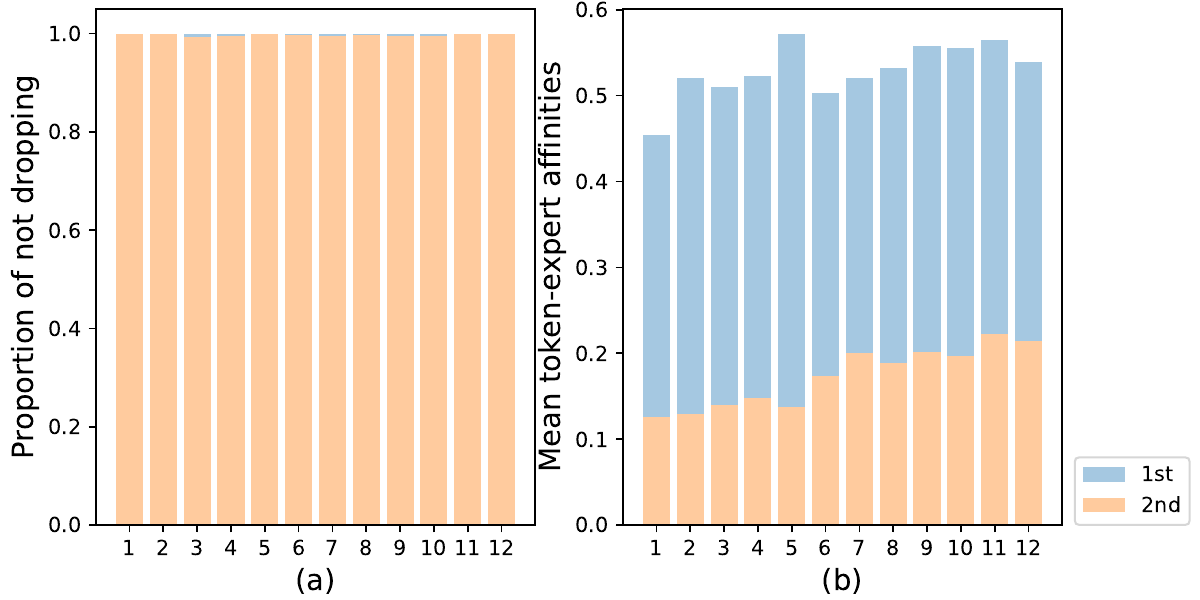} 
  \caption {\label{Fig7:TokenBalance}
    The proportion of not dropping and the mean token-expert affinities in top-2 routing are analyzed separately. 
    The data is derived from our MaxScore MoE with $e=16$ after training on $65$ billion tokens.
  }\vspace{-10pt}
\end{figure}

By comparing Figure~\ref{newFig2:ratio} and Figure~\ref{Fig7:TokenBalance}, we observe that network flow modeling effectively eliminates token dropping, and the $\mathrm{SoftTopk(\cdot)}$ operator significantly improves the distribution of token-expert affinities.

Our full model, GShard-SM (MaxScore), consistently achieves the best average performance of $43.44\%$, outperforming all ablated variants.
The synergistic combination of two methodological enhancements yields superadditive performance gains, with empirical results demonstrating that their integrated efficacy surpasses the linear summation of individual improvements.

\begin{table*}[t]
  \centering
  \scriptsize
  \renewcommand{\arraystretch}{1.4}
 \begin{tabularx}{0.94\linewidth}{lcccccccccccc} \toprule[1pt]
      \multirow{2}{*}{\textbf{Size}} & \multirow{2}{*}{\textbf{Model}} & ARC & ARC & \multirow{2}{*}{BoolQ} & Hella- & LAM- & \multirow{2}{*}{PIQA} & \multirow{2}{*}{RACE} & \multirow{2}{*}{SciQ} & \multirow{2}{*}{Record} & \multirow{2}{*}{OBQA} & \multirow{2}{*}{Avg.}  \\ 
     & & challenge & easy &  & Swag & BADA & & & & &  \\ \midrule[1pt]
    \multirow{3}{*}{\textbf{Base}}  & \textbf{GShard}   & $18.86$ & $\mathbf{44.49}$ & $61.90$ & $31.74$ & $21.54$ & $66.38$ & $\mathbf{28.52}$ & $69.4$ & $62.08$ & $16.2$ & $42.11$ \\
                                    & \textbf{DropLess} & $19.28$ & $44.07$ & $61.16$ & $32.03$ & $21.35$ & $67.14$ & $27.08$ & $67.9$ & $61.55$ & $16.0$ & $41.76$ \\
                                    & \textbf{MaxScore} & $\mathbf{20.73}$ & $\mathbf{44.49}$ & $\mathbf{62.23}$ & $\mathbf{32.85}$ & $\mathbf{23.27}$ & $\mathbf{67.41}$ & $\mathbf{28.52}$ & $\mathbf{72.5}$ & $\mathbf{64.00}$ & $\mathbf{18.4}$ & $\mathbf{43.44}$ \\ \hline
    \multirow{3}{*}{\textbf{Large}} & \textbf{GShard}   & $19.88$ & $45.58$ & $62.16$ & $33.34$ & $23.69$ & $67.74$ & $29.28$ & $70.0$ & $64.99$ & $19.2$ & $43.59$ \\
                                    & \textbf{DropLess} & $20.05$ & $45.24$ & $61.19$ & $33.97$ & $23.60$ & $67.63$ & $27.66$ & $69.7$ & $63.95$ & $17.2$ & $43.02$ \\
                                    & \textbf{MaxScore} & $\mathbf{20.90}$ & $\mathbf{45.92}$ & $\mathbf{62.39}$ & $\mathbf{34.00}$ & $\mathbf{24.96}$ & $\mathbf{68.28}$ & $\mathbf{29.67}$ & $\mathbf{74.3}$ & $\mathbf{66.12}$ & $\mathbf{19.8}$ & $\mathbf{44.63}$ \\ \hline
    \multirow{3}{*}{\textbf{XL}}    & \textbf{GShard}   & $20.05$ & $46.68$ & $63.09$ & $35.14$ & $25.05$ & $69.31$ & $29.19$ & $72.7$ & $67.50$ & $20.0$ & $44.87$ \\
                                    & \textbf{DropLess} & $20.14$ & $46.60$ & $61.69$ & $35.11$ & $24.34$ & $68.34$ & $29.19$ & $72.4$ & $67.55$ & $20.2$ & $44.56$ \\
                                    & \textbf{MaxScore} & $\mathbf{21.22}$ & $\mathbf{47.60}$ & $\mathbf{63.60}$ & $\mathbf{35.57}$ & $\mathbf{25.93}$ & $\mathbf{69.95}$ & $\mathbf{29.90}$ & $\mathbf{75.2}$ & $\mathbf{67.90}$ & $\mathbf{21.6}$ & $\mathbf{45.85}$ \\ \bottomrule[1pt]
\end{tabularx}
  \caption{\label{Tab6:Scaling}
    Results of scaling in model size.
  }\vspace{-5pt}
\end{table*}

\begin{table*}[t]
  \centering
  \scriptsize
  \renewcommand{\arraystretch}{1.4}
 \begin{tabularx}{0.95\linewidth}{lcccccccccccc} \toprule[1pt]
      \multirow{2}{*}{\textbf{Sparsity}} & \multirow{2}{*}{\textbf{Model}} & ARC & ARC & \multirow{2}{*}{BoolQ} & Hella- & LAM- & \multirow{2}{*}{PIQA} & \multirow{2}{*}{RACE} & \multirow{2}{*}{SciQ} & \multirow{2}{*}{Record} & \multirow{2}{*}{OBQA} & \multirow{2}{*}{Avg.}  \\ 
     & & challenge & easy &  & Swag & BADA & & & & &  \\ \midrule[1pt]
    \multirow{3}{*}{$\mathbf{2\!:\!16}$} & \textbf{GShard}   & $18.86$ & $\mathbf{44.49}$& $61.90$ & $31.74$ & $21.54$ & $66.38$ & $\mathbf{28.52}$ & $69.4$ & $62.08$ & $16.2$ & $42.11$ \\
                                         & \textbf{DropLess} & $19.28$ & $44.07$ & $61.16$ & $32.03$ & $21.35$ & $67.14$ & $27.08$ & $67.9$ & $61.55$ & $16.0$ & $41.76$ \\
                                         & \textbf{MaxScore} & $\mathbf{20.73}$ & $\mathbf{44.49}$ & $\mathbf{62.23}$ & $\mathbf{32.85}$ & $\mathbf{23.27}$ & $\mathbf{67.41}$ & $\mathbf{28.52}$ & $\mathbf{72.5}$ & $\mathbf{64.00}$ & $\mathbf{18.4}$ & $\mathbf{43.44}$ \\ \hline
    \multirow{3}{*}{$\mathbf{2\!:\!32}$} & \textbf{GShard}   & $19.62$ & $44.57$ & $62.28$ & $32.63$ & $21.99$ & $67.19$ & $\mathbf{29.04}$ & $69.6$ & $62.77$ & $18.2$ & $42.79$ \\
                                         & \textbf{DropLess} & $19.62$ & $44.51$ & $62.23$ & $32.36$ & $21.79$ & $67.10$ & $27.46$ & $68.3$ & $62.20$ & $16.8$ & $42.24$ \\
                                         & \textbf{MaxScore} & $\mathbf{20.90}$ & $\mathbf{44.60}$ & $\mathbf{63.73}$ & $\mathbf{33.24}$ & $\mathbf{23.76}$ & $\mathbf{67.63}$ & $\mathbf{29.04}$ & $\mathbf{73.5}$ & $\mathbf{64.41}$ & $\mathbf{18.8}$ & $\mathbf{43.96}$ \\ \hline
    \multirow{3}{*}{$\mathbf{2\!:\!64}$} & \textbf{GShard}   & $19.80$ & $44.86$ & $62.26$ & $33.05$ & $22.20$ & $67.30$ & $28.46$ & $69.4$ & $63.17$ & $17.6$ & $42.81$ \\
                                         & \textbf{DropLess} & $19.60$ & $44.69$ & $62.40$ & $32.79$ & $21.65$ & $67.27$ & $27.56$ & $69.5$ & $63.05$ & $17.6$ & $42.61$ \\
                                         & \textbf{MaxScore} & $\mathbf{21.11}$ & $\mathbf{46.17}$ & $\mathbf{64.24}$ & $\mathbf{33.38}$ & $\mathbf{23.41}$ & $\mathbf{67.95}$ & $\mathbf{28.90}$ & $\mathbf{73.3}$ & $\mathbf{64.60}$ & $\mathbf{19.0}$ & $\mathbf{44.21}$ \\ \bottomrule[1pt]
\end{tabularx}
  \caption{\label{Tab7:Sparsity}
    Results of scaling in sparsity.
  }\vspace{-5pt}
\end{table*}

\subsection{Scalability}

We perform scaling experiments along two dimensions: model size and sparsity. 
Detailed configurations are provided in Table~\ref{tab0:configs} and Table~\ref{tab5:ScalingExpert}. 

\begin{figure}[t]
  \centering
  \setlength{\abovecaptionskip}{4pt}
  \includegraphics[width=0.49\linewidth]{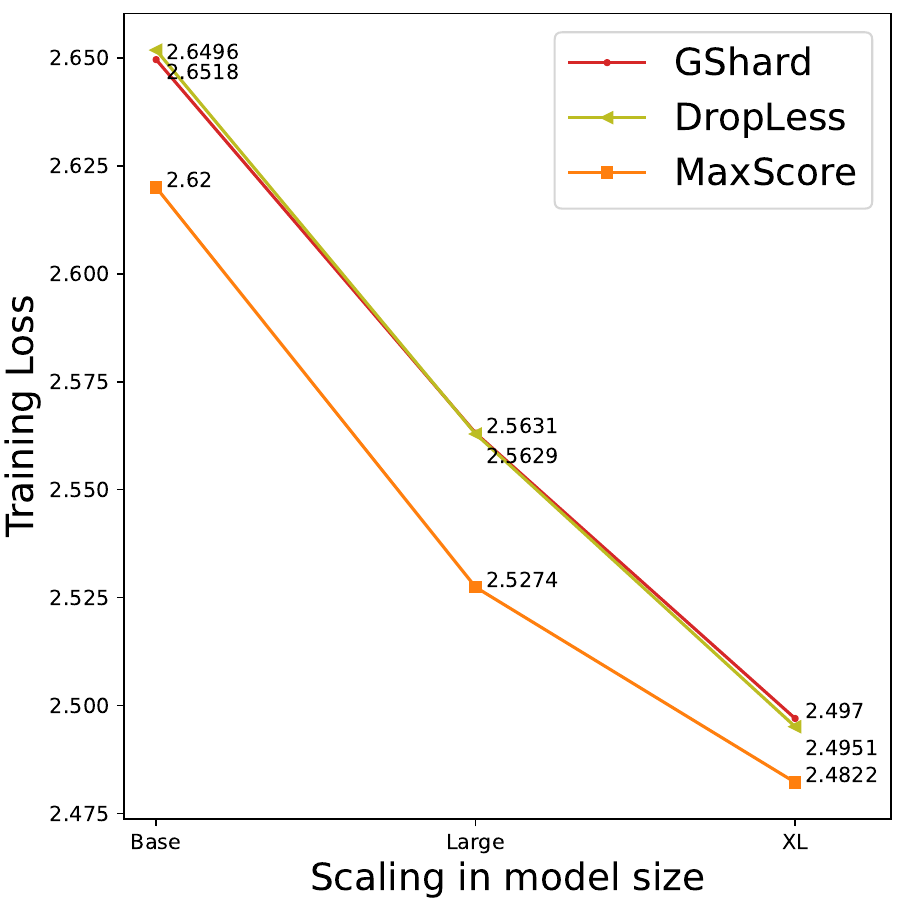} \hfill
  \includegraphics[width=0.49\linewidth]{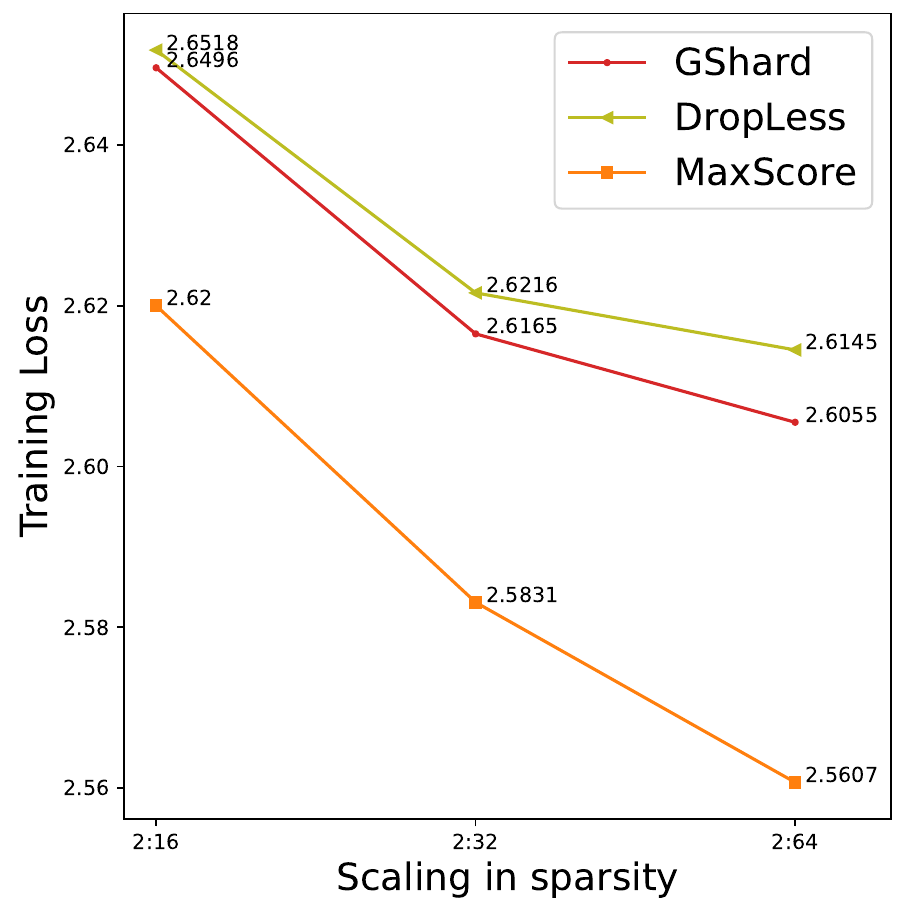}
  \caption {\label{Fig6:Scalability}
    Scalability with respect to model size and sparsity.
    The Y-axis represents the training loss of each model after training on approximately $65$ billion tokens.
  } \vspace{-12pt}
\end{figure}

As shown in Figure~\ref{Fig6:Scalability} and Tables~\ref{Tab6:Scaling} and~\ref{Tab7:Sparsity}, our MaxScore MoE consistently achieves a more significant reduction in training loss and superior evaluation performance compared to traditional MoE baselines such as GShard and DropLess across varying scales. 
In contrast, DropLess MoE suffers from increased expert load imbalance as sparsity increases, adversely affecting its scalability and overall performance. 
These results underscore MaxScore’s effectiveness in harnessing both model capacity and sparsity to improve MoE routing and model accuracy.

\begin{figure}[ht]
  \centering
  \setlength{\abovecaptionskip}{4pt}
  \includegraphics[width=\linewidth]{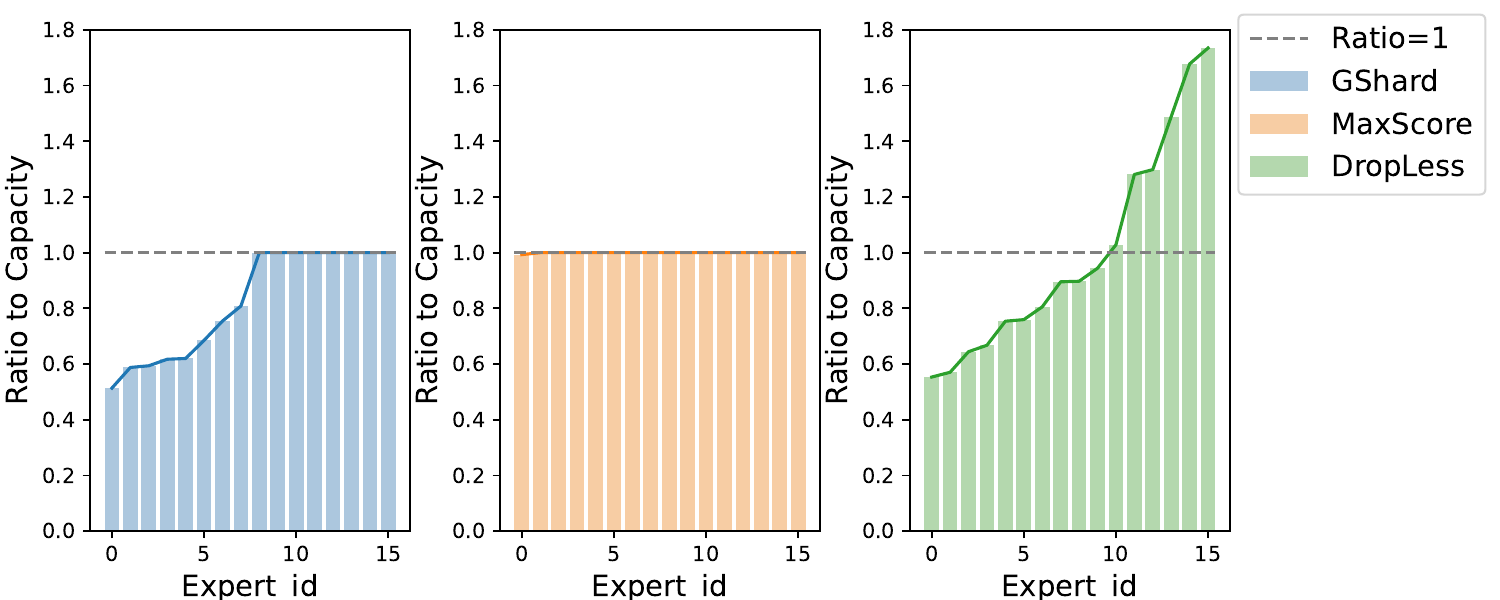} 
  \caption {\label{Fig8:ExpertBalance}
    The sorted ratio between the number of tokens each expert allocated and Capacity $c=k*n/e$ in the first layer of MoE with $k=2$ and $e=16$.
    For ExpertChoice MoE, the ratio is always equal to $1$.
    The mean ratios of GShard MoE, MaxScore MoE, and DropLess MoE are $0.8237$, $0.9996$, and $1$, respectively.
  }\vspace{-10pt}
\end{figure}

\subsection{Load Balancing Analysis}

Figure~\ref{Fig8:ExpertBalance} illustrates the sorted ratio between the number of tokens assigned to each expert and the capacity $c=\frac{k*n}{e}$ in the first MoE layer with $k=2$ and $e=16$ after training on about $65$ billion tokens.
For ExpertChoice MoE, this ratio remains strictly equal to $1$, indicating perfect load balancing by design.
MaxScore MoE achieves near-ideal load balance with a mean ratio of $0.9996$, closely approximating ExpertChoice.
In contrast, GShard exhibits notable load imbalance caused by token dropping, resulting in a lower mean ratio of $0.8237$ and uneven token distribution across experts.
DropLess displays extreme variability, with ratio values ranging from $0.55$ to $1.74$, indicating significant disparity in expert loads.
These findings demonstrate MaxScore’s superior capability in mitigating load imbalance relative to traditional approaches.

\subsection{Different SoftTopk Operators}

\begin{figure}[H]
  \centering
  \setlength{\abovecaptionskip}{4pt}
  \includegraphics[width=0.8\linewidth]{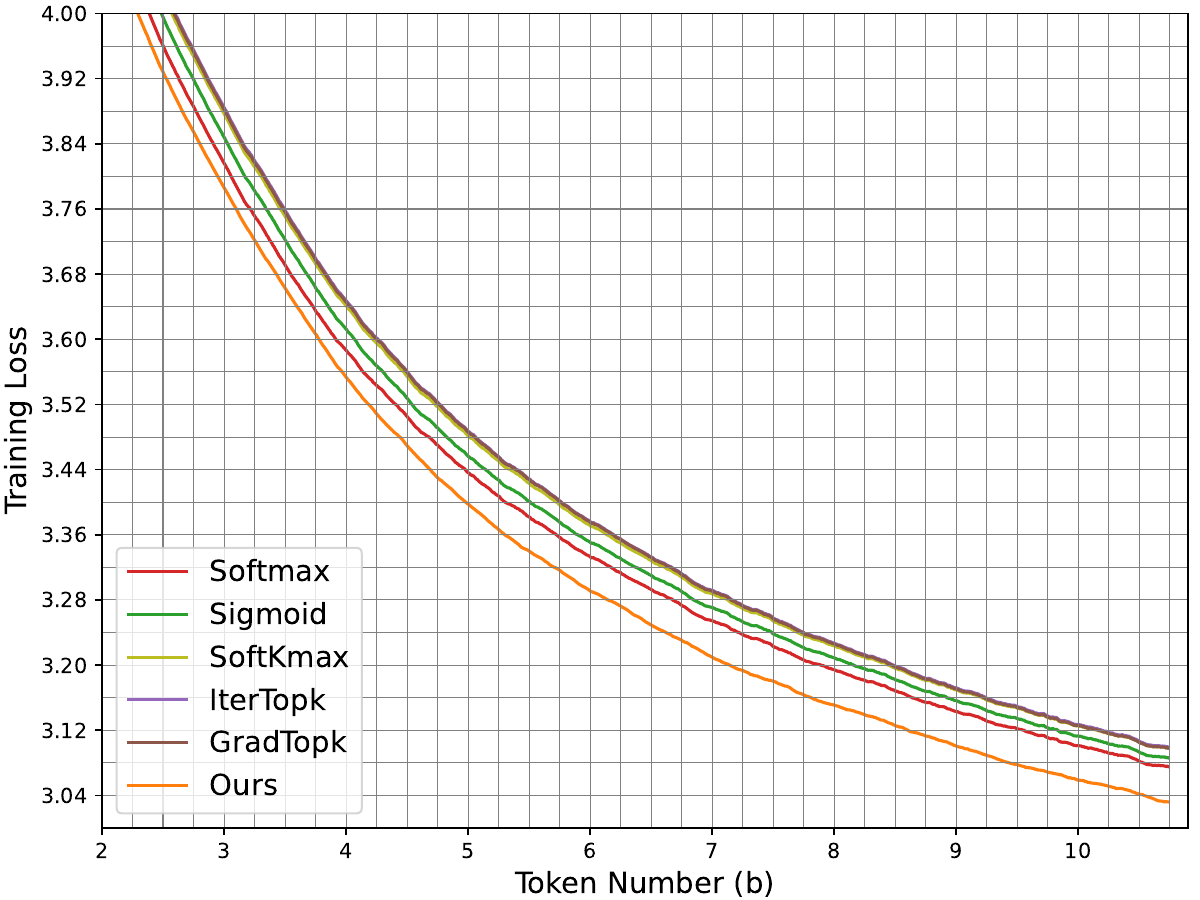} 
  \caption {\label{Fig9:SofttopK}
    Results of different operators.
  }\vspace{-10pt}
\end{figure}

We evaluate various $\mathrm{SoftTopk(\cdot)}$ operators listed in Table~\ref{Tab2:softtopk}. 
As illustrated in Figure~\ref{Fig9:SofttopK}, none yield performance improvements except for our proposed operator defined in Equation~(\ref{Eq:SoftTopk}). 
We hypothesize that the increased complexity of alternative operators may hinder effective model learning.

\subsection{Hyperparameter \textbf{t} in SoftTopk Operator}
\begin{figure}[H]
  \centering
  \setlength{\abovecaptionskip}{4pt}
  \includegraphics[width=0.8\linewidth]{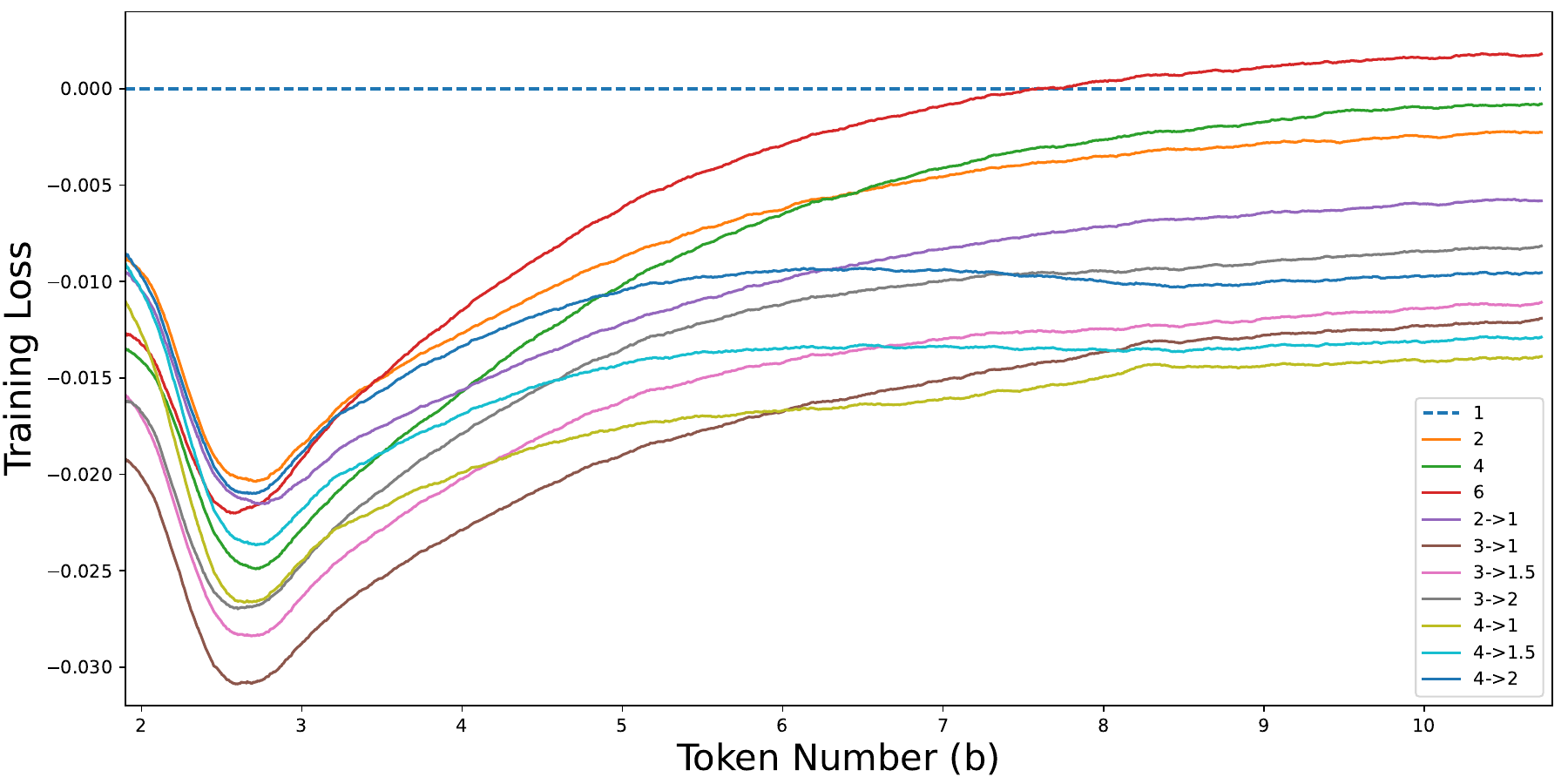} 
  \caption {\label{Fig10:hyperparameter}
    Hyperparameter tuning experiment.
  }
\end{figure}

We perform hyperparameter tuning for the parameter $\mathbf{t}$ in our $\mathrm{SoftTopk(\cdot)}$ operator defined in Equation~(\ref{Eq:SoftTopk}), exploring two strategies: maintaining a constant value or decaying t to 1 over training on 10b tokens. 
As shown in Figure~\ref{Fig10:hyperparameter}, the optimal approach initializes $\mathbf{t_0}$=4 and gradually decays it to 1.

\section{Conclusion and Future Work}

This work introduces MaxScore MoE, a novel mixture-of-experts routing paradigm formulated via minimum-cost maximum flow modeling and the integration of a differentiable $\mathrm{SoftTopk(\cdot)}$ operator.
To our knowledge, this is the first successful integration of network flow modeling and SoftTopk within MoE routing.
The synergistic combination of these components yields superadditive performance gains, with empirical evidence showing that their joint effect surpasses the linear sum of individual contributions.
Future work will focus on evaluating the method at larger model scales and across more diverse benchmarks to validate its generality and robustness.

\section*{Limitations}
Due to limited computational resources, our experiments are restricted to smaller-scale models, precluding direct comparison with larger, state-of-the-art models. 
Additionally, the training data volume is relatively modest; further experiments with substantially larger token budgets are necessary to fully assess the ultimate benefits and convergence properties of our approach.

\begin{table}[H]
  \centering
  \small
  \renewcommand{\arraystretch}{1.3}
  \begin{tabular}{lcccc} \toprule[1pt] \textbf{Model} & \textbf{Base} & \textbf{Large} & \textbf{XL} \\
    \midrule[1pt]
    \textbf{Activated Params} & $162M$ & $317M$ & $600M$ \\
    \textbf{Total Params} & $757M$ & $1.6B$ & $3.2B$ \\
    \textbf{FLOPs} & $302G$ & $603G$ & $1.2T$ \\ 
    \textbf{hidden\_size} & $768$ & $1128$ & $1608$ \\
    \textbf{num\_heads} & $12$ &  $12$ & $12$ \\
    \textbf{num\_layers} & $12$ &  $12$ & $12$ \\ 
    \bottomrule[1pt]
  \end{tabular}
  \caption{\label{tab0:configs}
    Configurations for different dense backbones.
    FLOPs are calculated with a single sequence of $512$ tokens.
    The $\mathrm{intermediate\_size}$ of the MLP layer in the dense model is four times that of the $\mathrm{hidden\_size}$, while for the top-$k$ MoE, the $\mathrm{intermediate\_size}$ in each expert is reduced to $1/k$, compared with the dense model.
  } \vspace{-10pt}
\end{table}

\begin{table}[H]
  \centering
  \small
  \renewcommand{\arraystretch}{1.3}
  \begin{tabular}{lccc}
    \toprule[1pt]
    \textbf{Sparsity} & $\mathbf{2\!:\!16}$ & $\mathbf{2\!:\!32}$ & $\mathbf{2\!:\!64}$ \\
    \midrule[1pt]
    \textbf{Activated Params} & $162M$ & $162M$ & $162M$ \\
    \textbf{Total Params} & $757M$ & $1475M$ & $2867M$ \\
    \textbf{FLOPs} & $302G$ & $302G$ & $302G$ \\ \bottomrule[1pt]
  \end{tabular}
  \caption{\label{tab5:ScalingExpert}
    Configurations of different sparsity.
  } \vspace{-10pt}
\end{table}

\begin{table}[H]
  \centering
  \small
  \renewcommand{\arraystretch}{1.3}
  \begin{tabular}{lcc} \toprule[1pt] \multirow{2}{*}{\textbf{Models}} & \textbf{Peak GPU}  & \textbf{Tokens processed} \\
  & \textbf{Memory Usage} & \textbf{Per Hour} \\ \midrule[1pt]
    \textbf{GShard} & $71.7 GB$ & $0.308b$ \\
    \textbf{ExpertChoice} & $71.7 GB$ & $0.301b$ \\
    \textbf{DropLess} & $73.4 GB$ & $0.296b$ \\
    \textbf{DeepSeek} & $\textbf{78.8 GB}$ & $\textbf{0.277b}$ \\
    \textbf{MaxScore-I} & $71.7 GB$ & $0.305b$ \\
    \textbf{GShard} & $71.7 GB$ & $0.299b$ \\
    \bottomrule[1pt]
  \end{tabular}
  \caption{\label{tab3:GPU}
    The peak GPU memory usage and the speed of processing of MoE models during training.
    DeepSeek MoE’s use of fine-grained experts leads to larger GPU memory and slower speed~\cite{DeepSeekV1,DeepSeekV2}. 
  } \vspace{-5pt}
\end{table}


\section*{Acknowledgments}
This work was supported in part by National Key Research and Development Program of China under Grant No. 2020YFA0804503, National Natural Science Foundation of China under Grant No. 62272264, and ByteDance Doubao Large Model Fund Project under Grant No. CT20240909109354. 



\bibliography{custom}

\begin{thebibliography}{47}
\providecommand{\natexlab}[1]{#1}

\bibitem[{Ainslie et~al.(2023)Ainslie, Lee-Thorp, de~Jong, Zemlyanskiy, Lebrón, and Sanghai}]{GQA}
Joshua Ainslie, James Lee-Thorp, Michiel de~Jong, Yury Zemlyanskiy, Federico Lebrón, and Sumit Sanghai. 2023.
\newblock \href {https://arxiv.org/abs/2305.13245} {Gqa: Training generalized multi-query transformer models from multi-head checkpoints}.
\newblock \emph{Preprint}, arXiv:2305.13245.

\bibitem[{Bellman(1958)}]{Bellman}
Richard Bellman. 1958.
\newblock On a routing problem.
\newblock \emph{Quarterly of applied mathematics}, 16(1):87--90.

\bibitem[{Bengio et~al.(2016)Bengio, Bacon, Pineau, and Precup}]{SoftConstraint}
Emmanuel Bengio, Pierre-Luc Bacon, Joelle Pineau, and Doina Precup. 2016.
\newblock \href {https://arxiv.org/abs/1511.06297} {Conditional computation in neural networks for faster models}.
\newblock \emph{Preprint}, arXiv:1511.06297.

\bibitem[{Bisk et~al.(2019)Bisk, Zellers, Bras, Gao, and Choi}]{PIQA}
Yonatan Bisk, Rowan Zellers, Ronan~Le Bras, Jianfeng Gao, and Yejin Choi. 2019.
\newblock \href {https://arxiv.org/abs/1911.11641} {Piqa: Reasoning about physical commonsense in natural language}.
\newblock \emph{Preprint}, arXiv:1911.11641.

\bibitem[{Chen et~al.(2016)Chen, Xu, Zhang, and Guestrin}]{activation_ckpt}
Tianqi Chen, Bing Xu, Chiyuan Zhang, and Carlos Guestrin. 2016.
\newblock \href {https://arxiv.org/abs/1604.06174} {Training deep nets with sublinear memory cost}.
\newblock \emph{Preprint}, arXiv:1604.06174.

\bibitem[{Clark et~al.(2022)Clark, de~las Casas, Guy, Mensch, Paganini, Hoffmann, Damoc, Hechtman, Cai, Borgeaud, van~den Driessche, Rutherford, Hennigan, Johnson, Millican, Cassirer, Jones, Buchatskaya, Budden, Sifre, Osindero, Vinyals, Rae, Elsen, Kavukcuoglu, and Simonyan}]{SBASE}
Aidan Clark, Diego de~las Casas, Aurelia Guy, Arthur Mensch, Michela Paganini, Jordan Hoffmann, Bogdan Damoc, Blake Hechtman, Trevor Cai, Sebastian Borgeaud, George van~den Driessche, Eliza Rutherford, Tom Hennigan, Matthew Johnson, Katie Millican, Albin Cassirer, Chris Jones, Elena Buchatskaya, David Budden, Laurent Sifre, Simon Osindero, Oriol Vinyals, Jack Rae, Erich Elsen, Koray Kavukcuoglu, and Karen Simonyan. 2022.
\newblock \href {https://arxiv.org/abs/2202.01169} {Unified scaling laws for routed language models}.
\newblock \emph{Preprint}, arXiv:2202.01169.

\bibitem[{Clark et~al.(2019)Clark, Lee, Chang, Kwiatkowski, Collins, and Toutanova}]{BoolQ}
Christopher Clark, Kenton Lee, Ming-Wei Chang, Tom Kwiatkowski, Michael Collins, and Kristina Toutanova. 2019.
\newblock \href {https://arxiv.org/abs/1905.10044} {Boolq: Exploring the surprising difficulty of natural yes/no questions}.
\newblock \emph{Preprint}, arXiv:1905.10044.

\bibitem[{Clark et~al.(2018)Clark, Cowhey, Etzioni, Khot, Sabharwal, Schoenick, and Tafjord}]{ARC}
Peter Clark, Isaac Cowhey, Oren Etzioni, Tushar Khot, Ashish Sabharwal, Carissa Schoenick, and Oyvind Tafjord. 2018.
\newblock \href {https://arxiv.org/abs/1803.05457} {Think you have solved question answering? try arc, the ai2 reasoning challenge}.
\newblock \emph{Preprint}, arXiv:1803.05457.

\bibitem[{Cuturi(2013)}]{Sinkhorn}
Marco Cuturi. 2013.
\newblock Sinkhorn distances: Lightspeed computation of optimal transport.
\newblock \emph{Advances in neural information processing systems}, 26.

\bibitem[{Dai et~al.(2024)Dai, Deng, Zhao, Xu, Gao, Chen, Li, Zeng, Yu, Wu, Xie, Li, Huang, Luo, Ruan, Sui, and Liang}]{DeepSeekV1}
Damai Dai, Chengqi Deng, Chenggang Zhao, R.~X. Xu, Huazuo Gao, Deli Chen, Jiashi Li, Wangding Zeng, Xingkai Yu, Y.~Wu, Zhenda Xie, Y.~K. Li, Panpan Huang, Fuli Luo, Chong Ruan, Zhifang Sui, and Wenfeng Liang. 2024.
\newblock \href {https://arxiv.org/abs/2401.06066} {Deepseekmoe: Towards ultimate expert specialization in mixture-of-experts language models}.
\newblock \emph{Preprint}, arXiv:2401.06066.

\bibitem[{DeepSeek-AI et~al.(2024{\natexlab{a}})DeepSeek-AI, Liu, Feng, Wang, Wang, Liu, Zhao, Dengr, Ruan, Dai, Guo, Yang, Chen, Ji, Li, Lin, Luo, Hao, Chen, Li, Zhang, Xu, Yang, Zhang, Ding, Xin, Gao, Li, Qu, Cai, Liang, Guo, Ni, Li, Chen, Yuan, Qiu, Song, Dong, Gao, Guan, Wang, Zhang, Xu, Xia, Zhao, Zhang, Li, Wang, Zhang, Zhang, Tang, Li, Tian, Huang, Wang, Zhang, Zhu, Chen, Du, Chen, Jin, Ge, Pan, Xu, Chen, Li, Lu, Zhou, Chen, Wu, Ye, Ma, Wang, Zhou, Yu, Zhou, Zheng, Wang, Pei, Yuan, Sun, Xiao, Zeng, An, Liu, Liang, Gao, Zhang, Li, Jin, Wang, Bi, Liu, Wang, Shen, Chen, Chen, Nie, Sun, Wang, Liu, Xie, Yu, Song, Zhou, Yang, Lu, Su, Wu, Li, Wei, Zhu, Xu, Huang, Li, Zhao, Sun, Li, Wang, Zheng, Zhang, Xiong, Zhao, He, Tang, Piao, Dong, Tan, Liu, Wang, Guo, Zhu, Wang, Zou, Zha, Ma, Yan, You, Liu, Ren, Ren, Sha, Fu, Huang, Zhang, Xie, Hao, Shao, Wen, Xu, Zhang, Li, Wang, Gu, Li, and Xie}]{DeepSeekV2}
DeepSeek-AI, Aixin Liu, Bei Feng, Bin Wang, Bingxuan Wang, Bo~Liu, Chenggang Zhao, Chengqi Dengr, Chong Ruan, Damai Dai, Daya Guo, Dejian Yang, Deli Chen, Dongjie Ji, Erhang Li, Fangyun Lin, Fuli Luo, Guangbo Hao, Guanting Chen, Guowei Li, H.~Zhang, Hanwei Xu, Hao Yang, Haowei Zhang, Honghui Ding, Huajian Xin, Huazuo Gao, Hui Li, Hui Qu, J.~L. Cai, Jian Liang, Jianzhong Guo, Jiaqi Ni, Jiashi Li, Jin Chen, Jingyang Yuan, Junjie Qiu, Junxiao Song, Kai Dong, Kaige Gao, Kang Guan, Lean Wang, Lecong Zhang, Lei Xu, Leyi Xia, Liang Zhao, Liyue Zhang, Meng Li, Miaojun Wang, Mingchuan Zhang, Minghua Zhang, Minghui Tang, Mingming Li, Ning Tian, Panpan Huang, Peiyi Wang, Peng Zhang, Qihao Zhu, Qinyu Chen, Qiushi Du, R.~J. Chen, R.~L. Jin, Ruiqi Ge, Ruizhe Pan, Runxin Xu, Ruyi Chen, S.~S. Li, Shanghao Lu, Shangyan Zhou, Shanhuang Chen, Shaoqing Wu, Shengfeng Ye, Shirong Ma, Shiyu Wang, Shuang Zhou, Shuiping Yu, Shunfeng Zhou, Size Zheng, T.~Wang, Tian Pei, Tian Yuan, Tianyu Sun, W.~L. Xiao, Wangding Zeng, Wei An, Wen
  Liu, Wenfeng Liang, Wenjun Gao, Wentao Zhang, X.~Q. Li, Xiangyue Jin, Xianzu Wang, Xiao Bi, Xiaodong Liu, Xiaohan Wang, Xiaojin Shen, Xiaokang Chen, Xiaosha Chen, Xiaotao Nie, Xiaowen Sun, Xiaoxiang Wang, Xin Liu, Xin Xie, Xingkai Yu, Xinnan Song, Xinyi Zhou, Xinyu Yang, Xuan Lu, Xuecheng Su, Y.~Wu, Y.~K. Li, Y.~X. Wei, Y.~X. Zhu, Yanhong Xu, Yanping Huang, Yao Li, Yao Zhao, Yaofeng Sun, Yaohui Li, Yaohui Wang, Yi~Zheng, Yichao Zhang, Yiliang Xiong, Yilong Zhao, Ying He, Ying Tang, Yishi Piao, Yixin Dong, Yixuan Tan, Yiyuan Liu, Yongji Wang, Yongqiang Guo, Yuchen Zhu, Yuduan Wang, Yuheng Zou, Yukun Zha, Yunxian Ma, Yuting Yan, Yuxiang You, Yuxuan Liu, Z.~Z. Ren, Zehui Ren, Zhangli Sha, Zhe Fu, Zhen Huang, Zhen Zhang, Zhenda Xie, Zhewen Hao, Zhihong Shao, Zhiniu Wen, Zhipeng Xu, Zhongyu Zhang, Zhuoshu Li, Zihan Wang, Zihui Gu, Zilin Li, and Ziwei Xie. 2024{\natexlab{a}}.
\newblock \href {https://arxiv.org/abs/2405.04434} {Deepseek-v2: A strong, economical, and efficient mixture-of-experts language model}.
\newblock \emph{Preprint}, arXiv:2405.04434.

\bibitem[{DeepSeek-AI et~al.(2024{\natexlab{b}})DeepSeek-AI, Liu, Feng, Xue, Wang, Wu, Lu, Zhao, Deng, Zhang, Ruan, Dai, Guo, Yang, Chen, Ji, Li, Lin, Dai, Luo, Hao, Chen, Li, Zhang, Bao, Xu, Wang, Zhang, Ding, Xin, Gao, Li, Qu, Cai, Liang, Guo, Ni, Li, Wang, Chen, Chen, Yuan, Qiu, Li, Song, Dong, Hu, Gao, Guan, Huang, Yu, Wang, Zhang, Xu, Xia, Zhao, Wang, Zhang, Li, Wang, Zhang, Zhang, Tang, Li, Tian, Huang, Wang, Zhang, Wang, Zhu, Chen, Du, Chen, Jin, Ge, Zhang, Pan, Wang, Xu, Zhang, Chen, Li, Lu, Zhou, Chen, Wu, Ye, Ye, Ma, Wang, Zhou, Yu, Zhou, Pan, Wang, Yun, Pei, Sun, Xiao, Zeng, Zhao, An, Liu, Liang, Gao, Yu, Zhang, Li, Jin, Wang, Bi, Liu, Wang, Shen, Chen, Zhang, Chen, Nie, Sun, Wang, Cheng, Liu, Xie, Liu, Yu, Song, Shan, Zhou, Yang, Li, Su, Lin, Li, Wang, Wei, Zhu, Zhang, Xu, Xu, Huang, Li, Zhao, Sun, Li, Wang, Yu, Zheng, Zhang, Shi, Xiong, He, Tang, Piao, Wang, Tan, Ma, Liu, Guo, Wu, Ou, Zhu, Wang, Gong, Zou, He, Zha, Xiong, Ma, Yan, Luo, You, Liu, Zhou, Wu, Ren, Ren, Sha, Fu, Xu, Huang, Zhang, Xie,
  Zhang, Hao, Gou, Ma, Yan, Shao, Xu, Wu, Zhang, Li, Gu, Zhu, Liu, Li, Xie, Song, Gao, and Pan}]{DeepSeekV3}
DeepSeek-AI, Aixin Liu, Bei Feng, Bing Xue, Bingxuan Wang, Bochao Wu, Chengda Lu, Chenggang Zhao, Chengqi Deng, Chenyu Zhang, Chong Ruan, Damai Dai, Daya Guo, Dejian Yang, Deli Chen, Dongjie Ji, Erhang Li, Fangyun Lin, Fucong Dai, Fuli Luo, Guangbo Hao, Guanting Chen, Guowei Li, H.~Zhang, Han Bao, Hanwei Xu, Haocheng Wang, Haowei Zhang, Honghui Ding, Huajian Xin, Huazuo Gao, Hui Li, Hui Qu, J.~L. Cai, Jian Liang, Jianzhong Guo, Jiaqi Ni, Jiashi Li, Jiawei Wang, Jin Chen, Jingchang Chen, Jingyang Yuan, Junjie Qiu, Junlong Li, Junxiao Song, Kai Dong, Kai Hu, Kaige Gao, Kang Guan, Kexin Huang, Kuai Yu, Lean Wang, Lecong Zhang, Lei Xu, Leyi Xia, Liang Zhao, Litong Wang, Liyue Zhang, Meng Li, Miaojun Wang, Mingchuan Zhang, Minghua Zhang, Minghui Tang, Mingming Li, Ning Tian, Panpan Huang, Peiyi Wang, Peng Zhang, Qiancheng Wang, Qihao Zhu, Qinyu Chen, Qiushi Du, R.~J. Chen, R.~L. Jin, Ruiqi Ge, Ruisong Zhang, Ruizhe Pan, Runji Wang, Runxin Xu, Ruoyu Zhang, Ruyi Chen, S.~S. Li, Shanghao Lu, Shangyan Zhou, Shanhuang
  Chen, Shaoqing Wu, Shengfeng Ye, Shengfeng Ye, Shirong Ma, Shiyu Wang, Shuang Zhou, Shuiping Yu, Shunfeng Zhou, Shuting Pan, T.~Wang, Tao Yun, Tian Pei, Tianyu Sun, W.~L. Xiao, Wangding Zeng, Wanjia Zhao, Wei An, Wen Liu, Wenfeng Liang, Wenjun Gao, Wenqin Yu, Wentao Zhang, X.~Q. Li, Xiangyue Jin, Xianzu Wang, Xiao Bi, Xiaodong Liu, Xiaohan Wang, Xiaojin Shen, Xiaokang Chen, Xiaokang Zhang, Xiaosha Chen, Xiaotao Nie, Xiaowen Sun, Xiaoxiang Wang, Xin Cheng, Xin Liu, Xin Xie, Xingchao Liu, Xingkai Yu, Xinnan Song, Xinxia Shan, Xinyi Zhou, Xinyu Yang, Xinyuan Li, Xuecheng Su, Xuheng Lin, Y.~K. Li, Y.~Q. Wang, Y.~X. Wei, Y.~X. Zhu, Yang Zhang, Yanhong Xu, Yanhong Xu, Yanping Huang, Yao Li, Yao Zhao, Yaofeng Sun, Yaohui Li, Yaohui Wang, Yi~Yu, Yi~Zheng, Yichao Zhang, Yifan Shi, Yiliang Xiong, Ying He, Ying Tang, Yishi Piao, Yisong Wang, Yixuan Tan, Yiyang Ma, Yiyuan Liu, Yongqiang Guo, Yu~Wu, Yuan Ou, Yuchen Zhu, Yuduan Wang, Yue Gong, Yuheng Zou, Yujia He, Yukun Zha, Yunfan Xiong, Yunxian Ma, Yuting Yan, Yuxiang
  Luo, Yuxiang You, Yuxuan Liu, Yuyang Zhou, Z.~F. Wu, Z.~Z. Ren, Zehui Ren, Zhangli Sha, Zhe Fu, Zhean Xu, Zhen Huang, Zhen Zhang, Zhenda Xie, Zhengyan Zhang, Zhewen Hao, Zhibin Gou, Zhicheng Ma, Zhigang Yan, Zhihong Shao, Zhipeng Xu, Zhiyu Wu, Zhongyu Zhang, Zhuoshu Li, Zihui Gu, Zijia Zhu, Zijun Liu, Zilin Li, Ziwei Xie, Ziyang Song, Ziyi Gao, and Zizheng Pan. 2024{\natexlab{b}}.
\newblock \href {https://arxiv.org/abs/2412.19437} {Deepseek-v3 technical report}.
\newblock \emph{Preprint}, arXiv:2412.19437.

\bibitem[{Eigen et~al.(2014)Eigen, Ranzato, and Sutskever}]{HardConstraint}
David Eigen, Marc'Aurelio Ranzato, and Ilya Sutskever. 2014.
\newblock \href {https://arxiv.org/abs/1312.4314} {Learning factored representations in a deep mixture of experts}.
\newblock \emph{Preprint}, arXiv:1312.4314.

\bibitem[{Fedus et~al.(2022)Fedus, Zoph, and Shazeer}]{SwitchTransformers}
William Fedus, Barret Zoph, and Noam Shazeer. 2022.
\newblock \href {https://arxiv.org/abs/2101.03961} {Switch transformers: Scaling to trillion parameter models with simple and efficient sparsity}.
\newblock \emph{Preprint}, arXiv:2101.03961.

\bibitem[{Ford(1956)}]{Ford}
Lester~Randolph Ford. 1956.
\newblock Network flow theory.
\newblock \emph{Rand Corporation Paper, Santa Monica, 1956}.

\bibitem[{Gale et~al.(2022)Gale, Narayanan, Young, and Zaharia}]{MegaBlocks}
Trevor Gale, Deepak Narayanan, Cliff Young, and Matei Zaharia. 2022.
\newblock \href {https://arxiv.org/abs/2211.15841} {Megablocks: Efficient sparse training with mixture-of-experts}.
\newblock \emph{Preprint}, arXiv:2211.15841.

\bibitem[{Gao et~al.(2024)Gao, Tow, Abbasi, Biderman, Black, DiPofi, Foster, Golding, Hsu, Le~Noac'h, Li, McDonell, Muennighoff, Ociepa, Phang, Reynolds, Schoelkopf, Skowron, Sutawika, Tang, Thite, Wang, Wang, and Zou}]{lm-eval}
Leo Gao, Jonathan Tow, Baber Abbasi, Stella Biderman, Sid Black, Anthony DiPofi, Charles Foster, Laurence Golding, Jeffrey Hsu, Alain Le~Noac'h, Haonan Li, Kyle McDonell, Niklas Muennighoff, Chris Ociepa, Jason Phang, Laria Reynolds, Hailey Schoelkopf, Aviya Skowron, Lintang Sutawika, Eric Tang, Anish Thite, Ben Wang, Kevin Wang, and Andy Zou. 2024.
\newblock \href {https://doi.org/10.5281/zenodo.12608602} {The language model evaluation harness}.

\bibitem[{Grattafiori et~al.(2024)Grattafiori, Dubey, Jauhri, Pandey, Kadian, Al-Dahle, Letman, Mathur, Schelten, Vaughan, Yang, Fan, Goyal, Hartshorn, Yang, Mitra, Sravankumar, Korenev, Hinsvark, Rao, Zhang, Rodriguez, Gregerson, Spataru, Roziere, Biron, Tang, Chern, Caucheteux, Nayak, Bi, Marra, McConnell, Keller, Touret, Wu, Wong, Ferrer, Nikolaidis, Allonsius, Song, Pintz, Livshits, Wyatt, Esiobu, Choudhary, Mahajan, Garcia-Olano, Perino, Hupkes, Lakomkin, AlBadawy, Lobanova, Dinan, Smith, Radenovic, Guzmán, Zhang, Synnaeve, Lee, Anderson, Thattai, Nail, Mialon, Pang, Cucurell, Nguyen, Korevaar, Xu, Touvron, Zarov, Ibarra, Kloumann, Misra, Evtimov, Zhang, Copet, Lee, Geffert, Vranes, Park, Mahadeokar, Shah, van~der Linde, Billock, Hong, Lee, Fu, Chi, Huang, Liu, Wang, Yu, Bitton, Spisak, Park, Rocca, Johnstun, Saxe, Jia, Alwala, Prasad, Upasani, Plawiak, Li, Heafield, Stone, El-Arini, Iyer, Malik, Chiu, Bhalla, Lakhotia, Rantala-Yeary, van~der Maaten, Chen, Tan, Jenkins, Martin, Madaan, Malo, Blecher,
  Landzaat, de~Oliveira, Muzzi, Pasupuleti, Singh, Paluri, Kardas, Tsimpoukelli, Oldham, Rita, Pavlova, Kambadur, Lewis, Si, Singh, Hassan, Goyal, Torabi, Bashlykov, Bogoychev, Chatterji, Zhang, Duchenne, Çelebi, Alrassy, Zhang, Li, Vasic, Weng, Bhargava, Dubal, Krishnan, Koura, Xu, He, Dong, Srinivasan, Ganapathy, Calderer, Cabral, Stojnic, Raileanu, Maheswari, Girdhar, Patel, Sauvestre, Polidoro, Sumbaly, Taylor, Silva, Hou, Wang, Hosseini, Chennabasappa, Singh, Bell, Kim, Edunov, Nie, Narang, Raparthy, Shen, Wan, Bhosale, Zhang, Vandenhende, Batra, Whitman, Sootla, Collot, Gururangan, Borodinsky, Herman, Fowler, Sheasha, Georgiou, Scialom, Speckbacher, Mihaylov, Xiao, Karn, Goswami, Gupta, Ramanathan, Kerkez, Gonguet, Do, Vogeti, Albiero, Petrovic, Chu, Xiong, Fu, Meers, Martinet, Wang, Wang, Tan, Xia, Xie, Jia, Wang, Goldschlag, Gaur, Babaei, Wen, Song, Zhang, Li, Mao, Coudert, Yan, Chen, Papakipos, Singh, Srivastava, Jain, Kelsey, Shajnfeld, Gangidi, Victoria, Goldstand, Menon, Sharma, Boesenberg,
  Baevski, Feinstein, Kallet, Sangani, Teo, Yunus, Lupu, Alvarado, Caples, Gu, Ho, Poulton, Ryan, Ramchandani, Dong, Franco, Goyal, Saraf, Chowdhury, Gabriel, Bharambe, Eisenman, Yazdan, James, Maurer, Leonhardi, Huang, Loyd, Paola, Paranjape, Liu, Wu, Ni, Hancock, Wasti, Spence, Stojkovic, Gamido, Montalvo, Parker, Burton, Mejia, Liu, Wang, Kim, Zhou, Hu, Chu, Cai, Tindal, Feichtenhofer, Gao, Civin, Beaty, Kreymer, Li, Adkins, Xu, Testuggine, David, Parikh, Liskovich, Foss, Wang, Le, Holland, Dowling, Jamil, Montgomery, Presani, Hahn, Wood, Le, Brinkman, Arcaute, Dunbar, Smothers, Sun, Kreuk, Tian, Kokkinos, Ozgenel, Caggioni, Kanayet, Seide, Florez, Schwarz, Badeer, Swee, Halpern, Herman, Sizov, Guangyi, Zhang, Lakshminarayanan, Inan, Shojanazeri, Zou, Wang, Zha, Habeeb, Rudolph, Suk, Aspegren, Goldman, Zhan, Damlaj, Molybog, Tufanov, Leontiadis, Veliche, Gat, Weissman, Geboski, Kohli, Lam, Asher, Gaya, Marcus, Tang, Chan, Zhen, Reizenstein, Teboul, Zhong, Jin, Yang, Cummings, Carvill, Shepard, McPhie,
  Torres, Ginsburg, Wang, Wu, U, Saxena, Khandelwal, Zand, Matosich, Veeraraghavan, Michelena, Li, Jagadeesh, Huang, Chawla, Huang, Chen, Garg, A, Silva, Bell, Zhang, Guo, Yu, Moshkovich, Wehrstedt, Khabsa, Avalani, Bhatt, Mankus, Hasson, Lennie, Reso, Groshev, Naumov, Lathi, Keneally, Liu, Seltzer, Valko, Restrepo, Patel, Vyatskov, Samvelyan, Clark, Macey, Wang, Hermoso, Metanat, Rastegari, Bansal, Santhanam, Parks, White, Bawa, Singhal, Egebo, Usunier, Mehta, Laptev, Dong, Cheng, Chernoguz, Hart, Salpekar, Kalinli, Kent, Parekh, Saab, Balaji, Rittner, Bontrager, Roux, Dollar, Zvyagina, Ratanchandani, Yuvraj, Liang, Alao, Rodriguez, Ayub, Murthy, Nayani, Mitra, Parthasarathy, Li, Hogan, Battey, Wang, Howes, Rinott, Mehta, Siby, Bondu, Datta, Chugh, Hunt, Dhillon, Sidorov, Pan, Mahajan, Verma, Yamamoto, Ramaswamy, Lindsay, Lindsay, Feng, Lin, Zha, Patil, Shankar, Zhang, Zhang, Wang, Agarwal, Sajuyigbe, Chintala, Max, Chen, Kehoe, Satterfield, Govindaprasad, Gupta, Deng, Cho, Virk, Subramanian, Choudhury,
  Goldman, Remez, Glaser, Best, Koehler, Robinson, Li, Zhang, Matthews, Chou, Shaked, Vontimitta, Ajayi, Montanez, Mohan, Kumar, Mangla, Ionescu, Poenaru, Mihailescu, Ivanov, Li, Wang, Jiang, Bouaziz, Constable, Tang, Wu, Wang, Wu, Gao, Kleinman, Chen, Hu, Jia, Qi, Li, Zhang, Zhang, Adi, Nam, Yu, Wang, Zhao, Hao, Qian, Li, He, Rait, DeVito, Rosnbrick, Wen, Yang, Zhao, and Ma}]{LLama3}
Aaron Grattafiori, Abhimanyu Dubey, Abhinav Jauhri, Abhinav Pandey, Abhishek Kadian, Ahmad Al-Dahle, Aiesha Letman, Akhil Mathur, Alan Schelten, Alex Vaughan, Amy Yang, Angela Fan, Anirudh Goyal, Anthony Hartshorn, Aobo Yang, Archi Mitra, Archie Sravankumar, Artem Korenev, Arthur Hinsvark, Arun Rao, Aston Zhang, Aurelien Rodriguez, Austen Gregerson, Ava Spataru, Baptiste Roziere, Bethany Biron, Binh Tang, Bobbie Chern, Charlotte Caucheteux, Chaya Nayak, Chloe Bi, Chris Marra, Chris McConnell, Christian Keller, Christophe Touret, Chunyang Wu, Corinne Wong, Cristian~Canton Ferrer, Cyrus Nikolaidis, Damien Allonsius, Daniel Song, Danielle Pintz, Danny Livshits, Danny Wyatt, David Esiobu, Dhruv Choudhary, Dhruv Mahajan, Diego Garcia-Olano, Diego Perino, Dieuwke Hupkes, Egor Lakomkin, Ehab AlBadawy, Elina Lobanova, Emily Dinan, Eric~Michael Smith, Filip Radenovic, Francisco Guzmán, Frank Zhang, Gabriel Synnaeve, Gabrielle Lee, Georgia~Lewis Anderson, Govind Thattai, Graeme Nail, Gregoire Mialon, Guan Pang,
  Guillem Cucurell, Hailey Nguyen, Hannah Korevaar, Hu~Xu, Hugo Touvron, Iliyan Zarov, Imanol~Arrieta Ibarra, Isabel Kloumann, Ishan Misra, Ivan Evtimov, Jack Zhang, Jade Copet, Jaewon Lee, Jan Geffert, Jana Vranes, Jason Park, Jay Mahadeokar, Jeet Shah, Jelmer van~der Linde, Jennifer Billock, Jenny Hong, Jenya Lee, Jeremy Fu, Jianfeng Chi, Jianyu Huang, Jiawen Liu, Jie Wang, Jiecao Yu, Joanna Bitton, Joe Spisak, Jongsoo Park, Joseph Rocca, Joshua Johnstun, Joshua Saxe, Junteng Jia, Kalyan~Vasuden Alwala, Karthik Prasad, Kartikeya Upasani, Kate Plawiak, Ke~Li, Kenneth Heafield, Kevin Stone, Khalid El-Arini, Krithika Iyer, Kshitiz Malik, Kuenley Chiu, Kunal Bhalla, Kushal Lakhotia, Lauren Rantala-Yeary, Laurens van~der Maaten, Lawrence Chen, Liang Tan, Liz Jenkins, Louis Martin, Lovish Madaan, Lubo Malo, Lukas Blecher, Lukas Landzaat, Luke de~Oliveira, Madeline Muzzi, Mahesh Pasupuleti, Mannat Singh, Manohar Paluri, Marcin Kardas, Maria Tsimpoukelli, Mathew Oldham, Mathieu Rita, Maya Pavlova, Melanie Kambadur,
  Mike Lewis, Min Si, Mitesh~Kumar Singh, Mona Hassan, Naman Goyal, Narjes Torabi, Nikolay Bashlykov, Nikolay Bogoychev, Niladri Chatterji, Ning Zhang, Olivier Duchenne, Onur Çelebi, Patrick Alrassy, Pengchuan Zhang, Pengwei Li, Petar Vasic, Peter Weng, Prajjwal Bhargava, Pratik Dubal, Praveen Krishnan, Punit~Singh Koura, Puxin Xu, Qing He, Qingxiao Dong, Ragavan Srinivasan, Raj Ganapathy, Ramon Calderer, Ricardo~Silveira Cabral, Robert Stojnic, Roberta Raileanu, Rohan Maheswari, Rohit Girdhar, Rohit Patel, Romain Sauvestre, Ronnie Polidoro, Roshan Sumbaly, Ross Taylor, Ruan Silva, Rui Hou, Rui Wang, Saghar Hosseini, Sahana Chennabasappa, Sanjay Singh, Sean Bell, Seohyun~Sonia Kim, Sergey Edunov, Shaoliang Nie, Sharan Narang, Sharath Raparthy, Sheng Shen, Shengye Wan, Shruti Bhosale, Shun Zhang, Simon Vandenhende, Soumya Batra, Spencer Whitman, Sten Sootla, Stephane Collot, Suchin Gururangan, Sydney Borodinsky, Tamar Herman, Tara Fowler, Tarek Sheasha, Thomas Georgiou, Thomas Scialom, Tobias Speckbacher,
  Todor Mihaylov, Tong Xiao, Ujjwal Karn, Vedanuj Goswami, Vibhor Gupta, Vignesh Ramanathan, Viktor Kerkez, Vincent Gonguet, Virginie Do, Vish Vogeti, Vítor Albiero, Vladan Petrovic, Weiwei Chu, Wenhan Xiong, Wenyin Fu, Whitney Meers, Xavier Martinet, Xiaodong Wang, Xiaofang Wang, Xiaoqing~Ellen Tan, Xide Xia, Xinfeng Xie, Xuchao Jia, Xuewei Wang, Yaelle Goldschlag, Yashesh Gaur, Yasmine Babaei, Yi~Wen, Yiwen Song, Yuchen Zhang, Yue Li, Yuning Mao, Zacharie~Delpierre Coudert, Zheng Yan, Zhengxing Chen, Zoe Papakipos, Aaditya Singh, Aayushi Srivastava, Abha Jain, Adam Kelsey, Adam Shajnfeld, Adithya Gangidi, Adolfo Victoria, Ahuva Goldstand, Ajay Menon, Ajay Sharma, Alex Boesenberg, Alexei Baevski, Allie Feinstein, Amanda Kallet, Amit Sangani, Amos Teo, Anam Yunus, Andrei Lupu, Andres Alvarado, Andrew Caples, Andrew Gu, Andrew Ho, Andrew Poulton, Andrew Ryan, Ankit Ramchandani, Annie Dong, Annie Franco, Anuj Goyal, Aparajita Saraf, Arkabandhu Chowdhury, Ashley Gabriel, Ashwin Bharambe, Assaf Eisenman, Azadeh
  Yazdan, Beau James, Ben Maurer, Benjamin Leonhardi, Bernie Huang, Beth Loyd, Beto~De Paola, Bhargavi Paranjape, Bing Liu, Bo~Wu, Boyu Ni, Braden Hancock, Bram Wasti, Brandon Spence, Brani Stojkovic, Brian Gamido, Britt Montalvo, Carl Parker, Carly Burton, Catalina Mejia, Ce~Liu, Changhan Wang, Changkyu Kim, Chao Zhou, Chester Hu, Ching-Hsiang Chu, Chris Cai, Chris Tindal, Christoph Feichtenhofer, Cynthia Gao, Damon Civin, Dana Beaty, Daniel Kreymer, Daniel Li, David Adkins, David Xu, Davide Testuggine, Delia David, Devi Parikh, Diana Liskovich, Didem Foss, Dingkang Wang, Duc Le, Dustin Holland, Edward Dowling, Eissa Jamil, Elaine Montgomery, Eleonora Presani, Emily Hahn, Emily Wood, Eric-Tuan Le, Erik Brinkman, Esteban Arcaute, Evan Dunbar, Evan Smothers, Fei Sun, Felix Kreuk, Feng Tian, Filippos Kokkinos, Firat Ozgenel, Francesco Caggioni, Frank Kanayet, Frank Seide, Gabriela~Medina Florez, Gabriella Schwarz, Gada Badeer, Georgia Swee, Gil Halpern, Grant Herman, Grigory Sizov, Guangyi, Zhang, Guna
  Lakshminarayanan, Hakan Inan, Hamid Shojanazeri, Han Zou, Hannah Wang, Hanwen Zha, Haroun Habeeb, Harrison Rudolph, Helen Suk, Henry Aspegren, Hunter Goldman, Hongyuan Zhan, Ibrahim Damlaj, Igor Molybog, Igor Tufanov, Ilias Leontiadis, Irina-Elena Veliche, Itai Gat, Jake Weissman, James Geboski, James Kohli, Janice Lam, Japhet Asher, Jean-Baptiste Gaya, Jeff Marcus, Jeff Tang, Jennifer Chan, Jenny Zhen, Jeremy Reizenstein, Jeremy Teboul, Jessica Zhong, Jian Jin, Jingyi Yang, Joe Cummings, Jon Carvill, Jon Shepard, Jonathan McPhie, Jonathan Torres, Josh Ginsburg, Junjie Wang, Kai Wu, Kam~Hou U, Karan Saxena, Kartikay Khandelwal, Katayoun Zand, Kathy Matosich, Kaushik Veeraraghavan, Kelly Michelena, Keqian Li, Kiran Jagadeesh, Kun Huang, Kunal Chawla, Kyle Huang, Lailin Chen, Lakshya Garg, Lavender A, Leandro Silva, Lee Bell, Lei Zhang, Liangpeng Guo, Licheng Yu, Liron Moshkovich, Luca Wehrstedt, Madian Khabsa, Manav Avalani, Manish Bhatt, Martynas Mankus, Matan Hasson, Matthew Lennie, Matthias Reso, Maxim
  Groshev, Maxim Naumov, Maya Lathi, Meghan Keneally, Miao Liu, Michael~L. Seltzer, Michal Valko, Michelle Restrepo, Mihir Patel, Mik Vyatskov, Mikayel Samvelyan, Mike Clark, Mike Macey, Mike Wang, Miquel~Jubert Hermoso, Mo~Metanat, Mohammad Rastegari, Munish Bansal, Nandhini Santhanam, Natascha Parks, Natasha White, Navyata Bawa, Nayan Singhal, Nick Egebo, Nicolas Usunier, Nikhil Mehta, Nikolay~Pavlovich Laptev, Ning Dong, Norman Cheng, Oleg Chernoguz, Olivia Hart, Omkar Salpekar, Ozlem Kalinli, Parkin Kent, Parth Parekh, Paul Saab, Pavan Balaji, Pedro Rittner, Philip Bontrager, Pierre Roux, Piotr Dollar, Polina Zvyagina, Prashant Ratanchandani, Pritish Yuvraj, Qian Liang, Rachad Alao, Rachel Rodriguez, Rafi Ayub, Raghotham Murthy, Raghu Nayani, Rahul Mitra, Rangaprabhu Parthasarathy, Raymond Li, Rebekkah Hogan, Robin Battey, Rocky Wang, Russ Howes, Ruty Rinott, Sachin Mehta, Sachin Siby, Sai~Jayesh Bondu, Samyak Datta, Sara Chugh, Sara Hunt, Sargun Dhillon, Sasha Sidorov, Satadru Pan, Saurabh Mahajan,
  Saurabh Verma, Seiji Yamamoto, Sharadh Ramaswamy, Shaun Lindsay, Shaun Lindsay, Sheng Feng, Shenghao Lin, Shengxin~Cindy Zha, Shishir Patil, Shiva Shankar, Shuqiang Zhang, Shuqiang Zhang, Sinong Wang, Sneha Agarwal, Soji Sajuyigbe, Soumith Chintala, Stephanie Max, Stephen Chen, Steve Kehoe, Steve Satterfield, Sudarshan Govindaprasad, Sumit Gupta, Summer Deng, Sungmin Cho, Sunny Virk, Suraj Subramanian, Sy~Choudhury, Sydney Goldman, Tal Remez, Tamar Glaser, Tamara Best, Thilo Koehler, Thomas Robinson, Tianhe Li, Tianjun Zhang, Tim Matthews, Timothy Chou, Tzook Shaked, Varun Vontimitta, Victoria Ajayi, Victoria Montanez, Vijai Mohan, Vinay~Satish Kumar, Vishal Mangla, Vlad Ionescu, Vlad Poenaru, Vlad~Tiberiu Mihailescu, Vladimir Ivanov, Wei Li, Wenchen Wang, Wenwen Jiang, Wes Bouaziz, Will Constable, Xiaocheng Tang, Xiaojian Wu, Xiaolan Wang, Xilun Wu, Xinbo Gao, Yaniv Kleinman, Yanjun Chen, Ye~Hu, Ye~Jia, Ye~Qi, Yenda Li, Yilin Zhang, Ying Zhang, Yossi Adi, Youngjin Nam, Yu, Wang, Yu~Zhao, Yuchen Hao, Yundi
  Qian, Yunlu Li, Yuzi He, Zach Rait, Zachary DeVito, Zef Rosnbrick, Zhaoduo Wen, Zhenyu Yang, Zhiwei Zhao, and Zhiyu Ma. 2024.
\newblock \href {https://arxiv.org/abs/2407.21783} {The llama 3 herd of models}.
\newblock \emph{Preprint}, arXiv:2407.21783.

\bibitem[{Hu et~al.(2024)Hu, Tu, Han, He, Cui, Long, Zheng, Fang, Huang, Zhao, Zhang, Thai, Zhang, Wang, Yao, Zhao, Zhou, Cai, Zhai, Ding, Jia, Zeng, Li, Liu, and Sun}]{WSD}
Shengding Hu, Yuge Tu, Xu~Han, Chaoqun He, Ganqu Cui, Xiang Long, Zhi Zheng, Yewei Fang, Yuxiang Huang, Weilin Zhao, Xinrong Zhang, Zheng~Leng Thai, Kaihuo Zhang, Chongyi Wang, Yuan Yao, Chenyang Zhao, Jie Zhou, Jie Cai, Zhongwu Zhai, Ning Ding, Chao Jia, Guoyang Zeng, Dahai Li, Zhiyuan Liu, and Maosong Sun. 2024.
\newblock \href {https://arxiv.org/abs/2404.06395} {Minicpm: Unveiling the potential of small language models with scalable training strategies}.
\newblock \emph{Preprint}, arXiv:2404.06395.

\bibitem[{Hwang et~al.(2023)Hwang, Cui, Xiong, Yang, Liu, Hu, Wang, Salas, Jose, Ram, Chau, Cheng, Yang, Yang, and Xiong}]{Tutel}
Changho Hwang, Wei Cui, Yifan Xiong, Ziyue Yang, Ze~Liu, Han Hu, Zilong Wang, Rafael Salas, Jithin Jose, Prabhat Ram, Joe Chau, Peng Cheng, Fan Yang, Mao Yang, and Yongqiang Xiong. 2023.
\newblock \href {https://arxiv.org/abs/2206.03382} {Tutel: Adaptive mixture-of-experts at scale}.
\newblock \emph{Preprint}, arXiv:2206.03382.

\bibitem[{Krajewski et~al.(2024)Krajewski, Ludziejewski, Adamczewski, Pióro, Krutul, Antoniak, Ciebiera, Król, Odrzygóźdź, Sankowski, Cygan, and Jaszczur}]{scalinglaw}
Jakub Krajewski, Jan Ludziejewski, Kamil Adamczewski, Maciej Pióro, Michał Krutul, Szymon Antoniak, Kamil Ciebiera, Krystian Król, Tomasz Odrzygóźdź, Piotr Sankowski, Marek Cygan, and Sebastian Jaszczur. 2024.
\newblock \href {https://arxiv.org/abs/2402.07871} {Scaling laws for fine-grained mixture of experts}.
\newblock \emph{Preprint}, arXiv:2402.07871.

\bibitem[{Lai et~al.(2017)Lai, Xie, Liu, Yang, and Hovy}]{RACE}
Guokun Lai, Qizhe Xie, Hanxiao Liu, Yiming Yang, and Eduard Hovy. 2017.
\newblock \href {https://arxiv.org/abs/1704.04683} {Race: Large-scale reading comprehension dataset from examinations}.
\newblock \emph{Preprint}, arXiv:1704.04683.

\bibitem[{Lepikhin et~al.(2020)Lepikhin, Lee, Xu, Chen, Firat, Huang, Krikun, Shazeer, and Chen}]{GShard}
Dmitry Lepikhin, HyoukJoong Lee, Yuanzhong Xu, Dehao Chen, Orhan Firat, Yanping Huang, Maxim Krikun, Noam Shazeer, and Zhifeng Chen. 2020.
\newblock \href {https://arxiv.org/abs/2006.16668} {Gshard: Scaling giant models with conditional computation and automatic sharding}.
\newblock \emph{Preprint}, arXiv:2006.16668.

\bibitem[{Loshchilov and Hutter(2019)}]{AdamW}
Ilya Loshchilov and Frank Hutter. 2019.
\newblock \href {https://arxiv.org/abs/1711.05101} {Decoupled weight decay regularization}.
\newblock \emph{Preprint}, arXiv:1711.05101.

\bibitem[{Martins and Astudillo(2016)}]{SparseMax}
André F.~T. Martins and Ramón~Fernandez Astudillo. 2016.
\newblock \href {https://arxiv.org/abs/1602.02068} {From softmax to sparsemax: A sparse model of attention and multi-label classification}.
\newblock \emph{Preprint}, arXiv:1602.02068.

\bibitem[{Mihaylov et~al.(2018)Mihaylov, Clark, Khot, and Sabharwal}]{OpenBookQA}
Todor Mihaylov, Peter Clark, Tushar Khot, and Ashish Sabharwal. 2018.
\newblock \href {https://arxiv.org/abs/1809.02789} {Can a suit of armor conduct electricity? a new dataset for open book question answering}.
\newblock \emph{Preprint}, arXiv:1809.02789.

\bibitem[{Muennighoff et~al.(2024)Muennighoff, Soldaini, Groeneveld, Lo, Morrison, Min, Shi, Walsh, Tafjord, Lambert, Gu, Arora, Bhagia, Schwenk, Wadden, Wettig, Hui, Dettmers, Kiela, Farhadi, Smith, Koh, Singh, and Hajishirzi}]{OLMoE}
Niklas Muennighoff, Luca Soldaini, Dirk Groeneveld, Kyle Lo, Jacob Morrison, Sewon Min, Weijia Shi, Pete Walsh, Oyvind Tafjord, Nathan Lambert, Yuling Gu, Shane Arora, Akshita Bhagia, Dustin Schwenk, David Wadden, Alexander Wettig, Binyuan Hui, Tim Dettmers, Douwe Kiela, Ali Farhadi, Noah~A. Smith, Pang~Wei Koh, Amanpreet Singh, and Hannaneh Hajishirzi. 2024.
\newblock \href {https://arxiv.org/abs/2409.02060} {Olmoe: Open mixture-of-experts language models}.
\newblock \emph{Preprint}, arXiv:2409.02060.

\bibitem[{Paperno et~al.(2016)Paperno, Kruszewski, Lazaridou, Pham, Bernardi, Pezzelle, Baroni, Boleda, and Fernández}]{LAMBADA}
Denis Paperno, Germán Kruszewski, Angeliki Lazaridou, Quan~Ngoc Pham, Raffaella Bernardi, Sandro Pezzelle, Marco Baroni, Gemma Boleda, and Raquel Fernández. 2016.
\newblock \href {https://arxiv.org/abs/1606.06031} {The lambada dataset: Word prediction requiring a broad discourse context}.
\newblock \emph{Preprint}, arXiv:1606.06031.

\bibitem[{Peters et~al.(2019)Peters, Niculae, and Martins}]{EntMax}
Ben Peters, Vlad Niculae, and André F.~T. Martins. 2019.
\newblock \href {https://arxiv.org/abs/1905.05702} {Sparse sequence-to-sequence models}.
\newblock \emph{Preprint}, arXiv:1905.05702.

\bibitem[{Raffel et~al.(2019)Raffel, Shazeer, Roberts, Lee, Narang, Matena, Zhou, Li, and Liu}]{2019t5}
Colin Raffel, Noam Shazeer, Adam Roberts, Katherine Lee, Sharan Narang, Michael Matena, Yanqi Zhou, Wei Li, and Peter~J. Liu. 2019.
\newblock \href {https://arxiv.org/abs/1910.10683} {Exploring the limits of transfer learning with a unified text-to-text transformer}.
\newblock \emph{arXiv e-prints}.

\bibitem[{Rajbhandari et~al.(2020)Rajbhandari, Rasley, Ruwase, and He}]{ZeRO}
Samyam Rajbhandari, Jeff Rasley, Olatunji Ruwase, and Yuxiong He. 2020.
\newblock \href {https://arxiv.org/abs/1910.02054} {Zero: Memory optimizations toward training trillion parameter models}.
\newblock \emph{Preprint}, arXiv:1910.02054.

\bibitem[{Shazeer(2020)}]{GLU}
Noam Shazeer. 2020.
\newblock \href {https://arxiv.org/abs/2002.05202} {Glu variants improve transformer}.
\newblock \emph{Preprint}, arXiv:2002.05202.

\bibitem[{Shazeer et~al.(2017)Shazeer, Mirhoseini, Maziarz, Davis, Le, Hinton, and Dean}]{MoE2017}
Noam Shazeer, Azalia Mirhoseini, Krzysztof Maziarz, Andy Davis, Quoc Le, Geoffrey Hinton, and Jeff Dean. 2017.
\newblock \href {https://arxiv.org/abs/1701.06538} {Outrageously large neural networks: The sparsely-gated mixture-of-experts layer}.
\newblock \emph{Preprint}, arXiv:1701.06538.

\bibitem[{Su(2024)}]{SoftTopk}
Jianlin Su. 2024.
\newblock \href {https://spaces.ac.cn/archives/10373} {After softmax: Finding a smooth approximation for top-k}.

\bibitem[{Su et~al.(2023)Su, Lu, Pan, Murtadha, Wen, and Liu}]{RoPE}
Jianlin Su, Yu~Lu, Shengfeng Pan, Ahmed Murtadha, Bo~Wen, and Yunfeng Liu. 2023.
\newblock \href {https://arxiv.org/abs/2104.09864} {Roformer: Enhanced transformer with rotary position embedding}.
\newblock \emph{Preprint}, arXiv:2104.09864.

\bibitem[{Touvron et~al.(2023{\natexlab{a}})Touvron, Lavril, Izacard, Martinet, Lachaux, Lacroix, Rozière, Goyal, Hambro, Azhar, Rodriguez, Joulin, Grave, and Lample}]{LLama}
Hugo Touvron, Thibaut Lavril, Gautier Izacard, Xavier Martinet, Marie-Anne Lachaux, Timothée Lacroix, Baptiste Rozière, Naman Goyal, Eric Hambro, Faisal Azhar, Aurelien Rodriguez, Armand Joulin, Edouard Grave, and Guillaume Lample. 2023{\natexlab{a}}.
\newblock \href {https://arxiv.org/abs/2302.13971} {Llama: Open and efficient foundation language models}.
\newblock \emph{Preprint}, arXiv:2302.13971.

\bibitem[{Touvron et~al.(2023{\natexlab{b}})Touvron, Martin, Stone, Albert, Almahairi, Babaei, Bashlykov, Batra, Bhargava, Bhosale, Bikel, Blecher, Ferrer, Chen, Cucurull, Esiobu, Fernandes, Fu, Fu, Fuller, Gao, Goswami, Goyal, Hartshorn, Hosseini, Hou, Inan, Kardas, Kerkez, Khabsa, Kloumann, Korenev, Koura, Lachaux, Lavril, Lee, Liskovich, Lu, Mao, Martinet, Mihaylov, Mishra, Molybog, Nie, Poulton, Reizenstein, Rungta, Saladi, Schelten, Silva, Smith, Subramanian, Tan, Tang, Taylor, Williams, Kuan, Xu, Yan, Zarov, Zhang, Fan, Kambadur, Narang, Rodriguez, Stojnic, Edunov, and Scialom}]{LLama2}
Hugo Touvron, Louis Martin, Kevin Stone, Peter Albert, Amjad Almahairi, Yasmine Babaei, Nikolay Bashlykov, Soumya Batra, Prajjwal Bhargava, Shruti Bhosale, Dan Bikel, Lukas Blecher, Cristian~Canton Ferrer, Moya Chen, Guillem Cucurull, David Esiobu, Jude Fernandes, Jeremy Fu, Wenyin Fu, Brian Fuller, Cynthia Gao, Vedanuj Goswami, Naman Goyal, Anthony Hartshorn, Saghar Hosseini, Rui Hou, Hakan Inan, Marcin Kardas, Viktor Kerkez, Madian Khabsa, Isabel Kloumann, Artem Korenev, Punit~Singh Koura, Marie-Anne Lachaux, Thibaut Lavril, Jenya Lee, Diana Liskovich, Yinghai Lu, Yuning Mao, Xavier Martinet, Todor Mihaylov, Pushkar Mishra, Igor Molybog, Yixin Nie, Andrew Poulton, Jeremy Reizenstein, Rashi Rungta, Kalyan Saladi, Alan Schelten, Ruan Silva, Eric~Michael Smith, Ranjan Subramanian, Xiaoqing~Ellen Tan, Binh Tang, Ross Taylor, Adina Williams, Jian~Xiang Kuan, Puxin Xu, Zheng Yan, Iliyan Zarov, Yuchen Zhang, Angela Fan, Melanie Kambadur, Sharan Narang, Aurelien Rodriguez, Robert Stojnic, Sergey Edunov, and Thomas
  Scialom. 2023{\natexlab{b}}.
\newblock \href {https://arxiv.org/abs/2307.09288} {Llama 2: Open foundation and fine-tuned chat models}.
\newblock \emph{Preprint}, arXiv:2307.09288.

\bibitem[{Vaswani et~al.(2017)Vaswani, Shazeer, Parmar, Uszkoreit, Jones, Gomez, Kaiser, and Polosukhin}]{AttentionIsAllYouNeed}
Ashish Vaswani, Noam Shazeer, Niki Parmar, Jakob Uszkoreit, Llion Jones, Aidan~N. Gomez, Lukasz Kaiser, and Illia Polosukhin. 2017.
\newblock \href {https://arxiv.org/abs/1706.03762} {Attention is all you need}.
\newblock \emph{CoRR}, abs/1706.03762.

\bibitem[{Waissi(1994)}]{NetworkFlow}
Gary~R Waissi. 1994.
\newblock Network flows: Theory, algorithms, and applications.

\bibitem[{Wang et~al.(2025)Wang, Zhu, and Chen}]{ReMoE}
Ziteng Wang, Jun Zhu, and Jianfei Chen. 2025.
\newblock \href {https://arxiv.org/abs/2412.14711} {Remoe: Fully differentiable mixture-of-experts with relu routing}.
\newblock \emph{Preprint}, arXiv:2412.14711.

\bibitem[{Welbl et~al.(2017)Welbl, Liu, and Gardner}]{SciQ}
Johannes Welbl, Nelson~F. Liu, and Matt Gardner. 2017.
\newblock \href {https://arxiv.org/abs/1707.06209} {Crowdsourcing multiple choice science questions}.
\newblock \emph{Preprint}, arXiv:1707.06209.

\bibitem[{Wolf et~al.(2020)Wolf, Debut, Sanh, Chaumond, Delangue, Moi, Cistac, Rault, Louf, Funtowicz, Davison, Shleifer, von Platen, Ma, Jernite, Plu, Xu, Scao, Gugger, Drame, Lhoest, and Rush}]{huggingface}
Thomas Wolf, Lysandre Debut, Victor Sanh, Julien Chaumond, Clement Delangue, Anthony Moi, Pierric Cistac, Tim Rault, Rémi Louf, Morgan Funtowicz, Joe Davison, Sam Shleifer, Patrick von Platen, Clara Ma, Yacine Jernite, Julien Plu, Canwen Xu, Teven~Le Scao, Sylvain Gugger, Mariama Drame, Quentin Lhoest, and Alexander~M. Rush. 2020.
\newblock \href {https://arxiv.org/abs/1910.03771} {Huggingface's transformers: State-of-the-art natural language processing}.
\newblock \emph{Preprint}, arXiv:1910.03771.

\bibitem[{Zellers et~al.(2019)Zellers, Holtzman, Bisk, Farhadi, and Choi}]{HellaSwag}
Rowan Zellers, Ari Holtzman, Yonatan Bisk, Ali Farhadi, and Yejin Choi. 2019.
\newblock \href {https://arxiv.org/abs/1905.07830} {Hellaswag: Can a machine really finish your sentence?}
\newblock \emph{Preprint}, arXiv:1905.07830.

\bibitem[{Zhang and Sennrich(2019)}]{RMSNorm}
Biao Zhang and Rico Sennrich. 2019.
\newblock \href {https://arxiv.org/abs/1910.07467} {Root mean square layer normalization}.
\newblock \emph{Preprint}, arXiv:1910.07467.

\bibitem[{Zhang et~al.(2018)Zhang, Liu, Liu, Gao, Duh, and Durme}]{ReCoRD}
Sheng Zhang, Xiaodong Liu, Jingjing Liu, Jianfeng Gao, Kevin Duh, and Benjamin~Van Durme. 2018.
\newblock \href {https://arxiv.org/abs/1810.12885} {Record: Bridging the gap between human and machine commonsense reading comprehension}.
\newblock \emph{Preprint}, arXiv:1810.12885.

\bibitem[{Zhou et~al.(2022)Zhou, Lei, Liu, Du, Huang, Zhao, Dai, Chen, Le, and Laudon}]{ExpertChoice}
Yanqi Zhou, Tao Lei, Hanxiao Liu, Nan Du, Yanping Huang, Vincent Zhao, Andrew Dai, Zhifeng Chen, Quoc Le, and James Laudon. 2022.
\newblock \href {https://arxiv.org/abs/2202.09368} {Mixture-of-experts with expert choice routing}.
\newblock \emph{Preprint}, arXiv:2202.09368.

\bibitem[{Zoph et~al.(2022)Zoph, Bello, Kumar, Du, Huang, Dean, Shazeer, and Fedus}]{Router-z-Loss}
Barret Zoph, Irwan Bello, Sameer Kumar, Nan Du, Yanping Huang, Jeff Dean, Noam Shazeer, and William Fedus. 2022.
\newblock \href {https://arxiv.org/abs/2202.08906} {St-moe: Designing stable and transferable sparse expert models}.
\newblock \emph{Preprint}, arXiv:2202.08906.

\end{thebibliography}

\appendix



\end{document}